\documentclass[journal,10pt]{IEEEtran}
\usepackage[utf8]{inputenc}

\usepackage{amsthm}
\usepackage{amsmath}
\usepackage{amssymb}
\usepackage{amsfonts}
\usepackage{cite}

\usepackage{booktabs}
\usepackage{bbm}
\usepackage{mathrsfs}

\newcommand{\Rmnum}[1]{\expandafter\@slowromancap\romannumeral #1@}
\usepackage{graphicx}
\usepackage{subfigure}
\usepackage{epsfig}
\usepackage[numbers,sort&compress]{natbib}
\usepackage{xcolor}
\usepackage{color}
\usepackage{enumerate}
\usepackage{extarrows}
\usepackage[ruled,linesnumbered,lined]{algorithm2e}  
\usepackage{algorithmic}
\usepackage{tensor}
\usepackage[colorlinks,linkcolor=blue,anchorcolor=blue,citecolor=blue,urlcolor=black]{hyperref}
\usepackage{url}

\DeclareMathOperator*{\argmax}{arg\,max}

\usepackage{makecell}
\usepackage{multirow}
\usepackage{diagbox}
\usepackage{subfigure}
\DeclareMathAlphabet\mathbfcal{OMS}{cmsy}{b}{n} 

\newtheorem{theorem}{Theorem}

\begin{document}

\title{Knowing When to Stop:\\ Delay-Adaptive Spiking Neural Network Classifiers with Reliability Guarantees}

\author{Jiechen Chen, \IEEEmembership{Member,~IEEE}, Sangwoo Park,  \IEEEmembership{Member,~IEEE}, Osvaldo Simeone,~\IEEEmembership{Fellow,~IEEE}
\vspace{-0.5cm} 
\thanks{The authors are with the King’s Communications, Learning and Information Processing (KCLIP) lab, King’s College London, London, WC2R 2LS, UK. (email:\{jiechen.chen, sangwoo.park, osvaldo.simeone\}@kcl.ac.uk). This work was supported by the European Union’s Horizon Europe project CENTRIC (101096379),  by~an Open Fellowship of the EPSRC (EP/W024101/1), and~by the EPSRC project (EP/X011852/1).}
}

\maketitle

 
\begin{abstract}
    Spiking neural networks (SNNs)  process time-series data via internal event-driven neural dynamics. The energy consumption of an SNN depends on the number of spikes exchanged between neurons over the course of the input presentation. Typically, decisions are produced after the entire input sequence has been processed. This results in latency and energy consumption levels that are fairly uniform across inputs. However, as explored in recent work, SNNs can produce an early decision when the SNN model is sufficiently ``confident'', adapting delay and energy consumption to the difficulty of each example. Existing techniques are based on heuristic measures of confidence that do not provide reliability guarantees, potentially exiting too early. In this paper, we introduce a novel delay-adaptive SNN-based inference methodology that, wrapping around any pre-trained SNN classifier, provides guaranteed reliability for the decisions produced at input-dependent stopping times. The approach, dubbed \emph{SpikeCP}, leverages tools from conformal prediction (CP). It  entails minimal complexity increase as compared to the underlying SNN, requiring only additional thresholding and counting operations at run time. SpikeCP is also extended to integrate a CP-aware training phase that targets delay performance. Variants of CP based on alternative confidence correction schemes, from Bonferroni to Simes, are explored, and extensive experiments are described using the MNIST-DVS data set, DVS128 Gesture dataset, and CIFAR-10 dataset.
\end{abstract}


\IEEEpeerreviewmaketitle
\newtheorem{fact}{Fact}
\newcommand{\mv}[1]{\mbox{\boldmath{$ #1 $}}}
\newcommand{\mb}[1]{\mathbb{#1}}
\newcommand{\Myfrac}[2]{\ensuremath{#1\mathord{\left/\right.\kern-\nulldelimiterspace}#2}}
\newcommand\Perms[2]{\tensor[^{#2}]P{_{#1}}}
\newcommand{\note}[1]{[\textcolor{red}{\textit{#1}}]}

\section{Introduction}
\label{Sec:intro}
\subsection{Motivation}
\begin{figure*}[htp]
	\centering
	\includegraphics[width=4.5in]{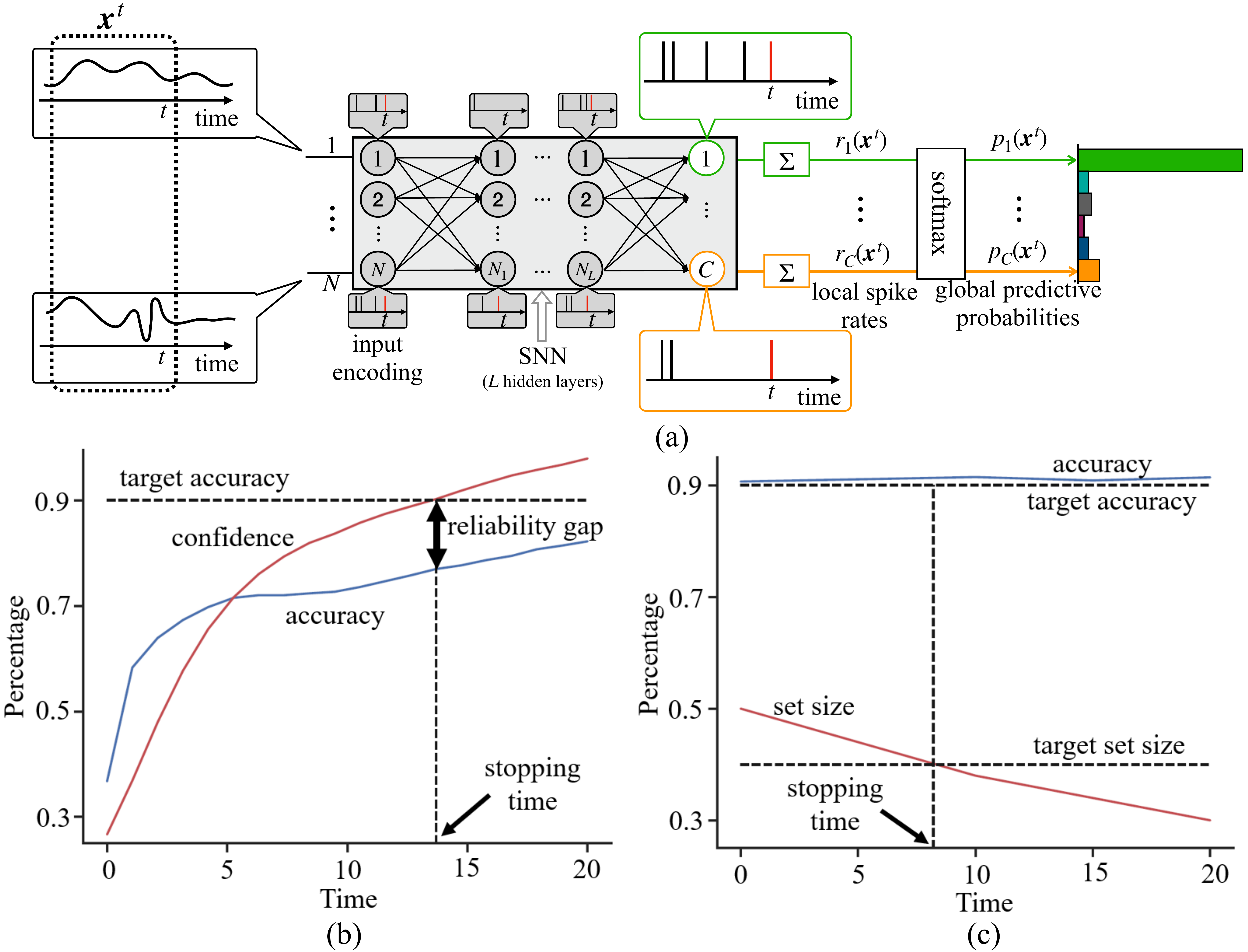}
    \caption{(a) SNN $C$-class classification model: At time $t$, real-valued discrete-time time-series data ${\mv x}^t$ are fed to the input neurons of an SNN and processed by internal spiking neurons, whose spikes feed $C$ readout neurons. Each output neuron $c \in \{1,...,C\}$ evaluates the \emph{local spike count} variable $r_c(\mv x^t)$ by accumulating the number of spikes it produces. The spike rates  may be aggregated across all output neurons to produce the \emph{predictive probability vector} $\{ p_c(\mv x^t)\}_{c=1}^C$. 
 (b) Evolution of confidence and accuracy as a function of time $t$ for a conventional pre-trained SNN. As illustrated, SNN classifiers tend to be first under-confident and then over-confident with respect to the true accuracy, which may cause a positive \emph{reliability gap}, i.e., a shortfall in accuracy, when the confidence level is used as an inference-stopping criterion. (c) Evolution of the (test-averaged) predicted set size (normalized by the number of classes $C=10$) and of the set accuracy as a function of time $t$ for the same pre-trained SNN when used in conjunction with the proposed SpikeCP method. The set accuracy is the probability that the true label lies inside the predicted set. It is observed that, irrespective of the stopping time, the set accuracy is always guaranteed to exceed the target accuracy level. Therefore, the inference-stopping criterion can be designed to control the trade-off between latency, and hence also energy consumption, and the size of the predicted set. }
	\label{fig:overall}
 \vspace{-0.5cm}
\end{figure*}

Spiking neural networks (SNNs) have emerged as efficient models for the processing of time series data, particularly in settings characterized by sparse inputs \cite{davies2018loihi}. SNNs implement recurrent,  event-driven, neural dynamics whose energy consumption depends on the number of spikes exchanged between neurons over the course of the input presentation. As shown in Fig.~\ref{fig:overall}(a), an SNN-based classifier processes input time series to produce spiking signals -- one for each possible class -- with the spiking rate of each output signal typically quantifying the \emph{confidence} the model has in the corresponding labels. Typically, decisions are produced after the entire input sequence has been processed, resulting in latency and energy consumption levels that are fairly uniform across inputs. 

The online operation of SNNs, along with their in-built adaptive measures of confidence derived from the output spikes, suggest an alternative operating principle, whereby inference latency and energy consumption are tailored to the difficulty of each example. Specifically, as proposed in \cite{li2023unleashing, li2023seenn}, \emph{delay-adaptive} SNN classifiers  produce an \emph{early decision} when the SNN model is sufficiently confident. In practice, however, the confidence levels output by an SNN, even when adjusted with limited data as in \cite{li2021free}, are not well \emph{calibrated}, in the sense that they do not precisely reflect the underlying accuracy of the corresponding decisions (see Fig. \ref{fig:overall}). As a result, relying on its output confidence signals may cause the SNN to stop prematurely, failing to meet target accuracy  levels.

To illustrate this problem, Fig.~\ref{fig:overall}(b)  shows the test accuracy and confidence level (averaged over test inputs) that are produced by a pre-trained SNN for an image classification task (on the MNIST-DVS dataset \cite{serrano2015poker}) as a function of time $t$. It is observed that the SNN's classification decisions  tend to be first \emph{under-confident} and then \emph{over-confident} with respect to the decision's ground-truth, unknown, test accuracy. Therefore, using the SNN's confidence levels to decide when to make a decision generally causes a \emph{reliability gap} between the true test accuracy and the target accuracy. This problem can be mitigated by relying on \emph{calibration data} to re-calibrate the SNN's confidence level, but only if one has enough calibration data \cite{li2021free} (see Sec. \ref{exp} for experimental evidence, e.g., in Fig. \ref{r1}).

\subsection{SpikeCP}
In this paper, we introduce a novel delay-adaptive SNN solution that (\emph{i}) provides \emph{guaranteed} reliability -- and hence a zero (or non-positive) reliability gap; while (\emph{ii}) supporting a tunable trade-off between latency and inference energy, on the one hand, and informativeness of the decision, on the other hand. The proposed method, referred to as \emph{SpikeCP},  builds on \emph{conformal prediction} (CP), a statistical framework for calibration that is currently experiencing a surge of interest in the machine learning community  \cite{vovk2022algorithmic, angelopoulos2107gentle}.

SpikeCP uses local or global information produced by the output layer of SNN model (see Fig.~\ref{fig:overall}(a)), along with \emph{calibration data}, to produce at each time $t$ a \emph{subset of labels} as its decision. By the properties of CP, the predictive set produced by SpikeCP includes the ground-truth label with a target accuracy level at any stopping time. A stopping decision is then made by SpikeCP not based on a reliability requirement -- which is always satisfied -- but rather based on the desired size of the predicted set. As illustrated in Fig.~\ref{fig:overall}(c), the desired set size provides a novel degree of freedom that can be used to control the trade-off between latency, or energy consumption, and \emph{informativeness} of the decision, as measured by the set size. 

SpikeCP wraps around any pre-trained SNN classifier, providing guaranteed reliability for the decisions produced at input-dependent stopping times. It does so with a minimal complexity increase as compared to the underlying SNN, requiring only additional thresholding and counting operations at run time. At a technical level, SpikeCP applies a Bonferroni correction of the target accuracy that scales with the number of possible stopping times in order to ensure a zero (or non-positive) reliability gap. Heuristics based on Simes correction \cite{fisch2020efficient,rodland2006simes} are also explored via numerical results.

The approach is finally extended to integrate a CP-aware training phase that targets minimization of the delay via a reduction of the average predicted set size. Unlike conventional training methods for SNNs \cite{wade2010swat}, the proposed method adds an explicit regularizer that controls the average number of labels included in the predicted set.

\subsection{Related Work}
\label{sec:related_work}
\emph{Training SNNs.} Typical training algorithms for SNNs are based on direct conversions from trained artificial neural networks \cite{li2023unleashing}, on heuristic local rules such as spike-timing-dependent plasticity (STDP) \cite{caporale2008spike}, or approximations of backpropagation-through-time that simplify credit assignment and address the non-differentiability of the spiking mechanism \cite{neftci2019surrogate, davies2018loihi}. Another approach that targets the direct training of SNNs is based on  modelling the spiking mechanism as a stochastic process, which enables the use of likelihood-based methods \cite{jang2019introduction}, as well as of Bayesian rules \cite{skatchkovsky2022bayesian}. SpikeCP works as a wrapper around any training scheme.

\emph{Calibration and delay-adaptivity for SNN.} Calibration is a subject of extensive research for artificial neural networks \cite{guo2017calibration, quach2023conformal} but is still an underexplored subject for SNNs.  SNN calibration is carried out by leveraging a pre-trained ANN in \cite{li2021free}; while \cite{skatchkovsky2022bayesian} applies Bayesian learning to reduce the calibration error. As discussed in the previous sections,  adaptivity for rate decoding 
 was studied in \cite{li2023unleashing, li2023seenn}. Other forms of  adaptivity may leverage \emph{temporal decoding}, whereby, for instance, as soon as one output neuron spikes a decision is made \cite{rosenfeld2019learning}. 

\emph{Early exit in conventional deep learning.} The idea of delay-adaptivity in SNNs is related to that of \emph{early-exit} decisions in feedforward neural networks. In neural networks with an early exit option, confidence levels are evaluated at intermediate layers, and a decision is made when the confidence level passes a threshold \cite{panda2016conditional,teerapittayanon2016branchynet,laskaridis2020spinn}. The role of calibration for early-exit neural networks was studied in \cite{pacheco2023impact}.

\emph{Prediction cascades.} Another related concept is that of prediction cascades, which apply a sequence of classifiers, ranging from light-weight to computationally expensive \cite{weiss2010structured}, to a static input. The goal is to apply the more expensive classifiers only when the difficulty of the input requires it. The application of CP to prediction cascades was investigated in \cite{fisch2020efficient}. 

\emph{CP-aware training.} CP provides a general methodology to turn a pre-trained probabilistic predictors into a reliable set predictor \cite{angelopoulos2107gentle,cohen2022calibrating}. Applications of CP range from healthcare \cite{lin2022conformal} to control \cite{angelopoulos2022conformal}, large language models \cite{kumar2023conformal}, and wireless systems \cite{cohen2022calibrating}.  References \cite{stutz2021learning, einbinder2022training, park2022few} have observed that the efficiency of the set-valued predictions produced via CP can be improved by training the underlying predictor in a \emph{CP-aware} manner that targets directly the predicted set size. Specifically, the authors of \cite{stutz2021learning} propose to minimize a loss functions that penalizes large prediction set sizes when used in conjunction with CP. It explored strategies to differentiate through CP during training with the goal of training model,  with the conformal wrapper end-to-end.  Related work in \cite{park2022few} has leveraged differentiation through CP to design meta-learning strategies targeting the predictive set size (see also \cite{fisch2021few}). To the best of our knowledge, no prior work has applied the idea of CP-aware training to the design of delay-adaptive classifiers. 

\subsection{Main Contributions and Paper Organization}
The main contributions of this paper are summarized as follows. \\
\noindent $\bullet$ We introduce \emph{SpikeCP}, a novel inference framework that turns any pre-trained SNN into a reliable and delay-adaptive set predictor, irrespective of the quality of the pre-trained SNN and of the number of calibration points. The performance of the pre-trained SNN determines the achievable trade-off curve between latency and energy efficiency, on the one hand, and informativeness of the decision, as measured by the set size, on the other. SpikeCP requires minimal changes to the underlying SNN, adding only counting and thresholding operations. Furthermore, it can be implemented using different measures of confidence at the output of the SNN, such as spiking rates and softmax-modulated signals. \\
\noindent $\bullet$ \emph{Theoretical guarantees} are proved by leveraging a modification of the confidence levels based on Bonferroni correction \cite{vovk2022admissible}. Heuristic alternatives based on Simes correction are also considered \cite{vovk2021values}. \\
\noindent $\bullet$ In order to improve the performance in terms of attainable trade-offs between delay/energy consumption and predictive set sizes, we introduce a \emph{SpikeCP-aware training} strategy that targets directly the performance of the SNN when used in conjunction with SpikeCP. The approach is based on regularizing the classical cross-entropy loss \cite{9997098, skatchkovsky2021spiking} with a differentiable approximation of the predicted set size.\\
\noindent $\bullet$ Extensive numerical results are provided that demonstrate the advantages of the proposed SpikeCP algorithms over conventional point predictors in terms of reliability, latency, and energy consumption metrics.

The remainder of the paper is organized as follows. Section \ref{problem definition} presents the multi-class classification problem via SNNs. Adaptive point classification schemes are reviewed for reference in Section \ref{sec:adaptive_point_classification}. The SpikeCP algorithm is proposed in Section \ref{spikecp}, while Section \ref{spikecp+} presents a training strategy that targets directly the performance of the SNN when used in conjunction with SpikeCP. Experimental setting and results are described in Section \ref{exp}. Finally, Section \ref{conclusion} concludes the paper.

\section{Problem Definition} \label{problem definition}
In this paper, we consider the problem of efficiently and reliably classifying time series data via SNNs by integrating adaptive-latency decision rules \cite{li2023unleashing, li2023seenn} with CP \cite{vovk2022algorithmic, angelopoulos2107gentle}. The proposed scheme, SpikeCP, produces \emph{adaptive SNN-based set classifiers} with \emph{formal reliability guarantees}.  In this section, we start by defining the problem under study, along with the main performance metrics of interest, namely reliability, latency, and inference energy. We also review the conventional model of SNNs adopted in this study that is based on leaky integrate-and-fire (LIF) neurons  \cite{eshraghian2021training}.

\subsection{Multi-Class Time Series Classification} \label{classp}
We focus on the problem of classifying real-valued vector time series data $\mv x=(\mv x_1,..., \mv x_T)$, with $N \times 1$ vector samples $\mv x_t$ over time index $t=1,...,T$,  into $C$ classes, using \emph{dynamic} classifiers implemented via SNNs. As illustrated in Fig. \ref{fig:overall}(a), the SNN model has $N$ input neurons, an arbitrary number of internal spiking neurons, and $C$ output neurons in the readout layer. Each output neuron is associated with one of the $C$ class labels in set $\mathcal{C}=\{1,...,C\}$.

At each time $t$, the SNN takes as input the real-valued vector $\mv x_t$, and produces sequentially the binary, ``spiking'', output vector $\mv y_t=[y_{t,1}, ..., y_{t,C}]$ of size $C$, with $y_{t,c}\in \{0,1\}$, as a function of the samples 
\begin{align}\label{eq:firstsamples}
    \mv x^t = (\mv x_1, ..., \mv x_t),
\end{align}
observed so far. Accordingly, if $y_{t,c}=1$, output neuron $c\in\mathcal{C}$ emits a spike, while, if $y_{t,c}=0$, output neuron $c$ is silent. Using conventional \emph{rate decoding}, each output neuron $c\in\mathcal{C}$ maintains the sum of spikes evaluated so far, i.e., 
\begin{equation}
    r_c(\mv x^t)= \sum_{t^{\prime}=1}^{t} y_{t^{\prime}, c}, \label{count}
\end{equation}
along the time axis $t=1,...,T$.

Each \emph{spike count} variable $r_c(\mv x^t)$ may be used as an estimate of the degree of confidence of the SNN in class $c$ being the correct one. In order to obtain predictive probabilities, the spike count vector $\mathbf{r}(\mv x^t)=[r_1(\mv x^t), ..., r_C(\mv x^t)]$ can be passed through a softmax function to yield a probability for class $c$ as $p_c(\mv x^t) = e^{r_c(\scalebox{0.7}{$\mv x^t$})}/{\sum_{c^{\prime}=1}^C e^{r_{c^{\prime}}(\scalebox{0.7}{$\mv x^t$})}}$ (see Fig. \ref{fig:overall}(a)). The resulting \emph{predictive probability vector} 
\begin{equation} 
\mathbf{p}(\mv x^t)=[p_1(\mv x^t), ..., p_C(\mv x^t)], \label{prob}
\end{equation} 
quantifies the \emph{normalized} confidence levels of the classifier in each class $c$ given the observations up to time $t$. We emphasize that evaluating the vector \eqref{prob} requires coordination among all output neurons, since each probability value $p_c(\mv x^t)$ depends on the spike counts of all output spiking neurons. 

A classifier is said to be \emph{well calibrated} if the confidence vector $\mathbf{p}(\mv x^t)$ provides a close approximation of the true, test, accuracy of each decision $c \in \mathcal{C}$. Machine learning models based on deep learning are well known to be typically  \emph{over-confident}, resulting in confidence vectors  $\mathbf{p}(\mv x^t)$ that are excessively skewed towards a single class $c$, dependent on the input $\mv x^t$ \cite{guo2017calibration,tao2023}. As discussed in Sec.~\ref{Sec:intro}, SNN models also tend to provide over-confident decisions as time $t$ increases.

Following the conventional supervised learning formulation of the problem, multi-class time series classification data consist of pairs $(\mv x, c)$ of input sequence $\mv x$ and true class index $c\in\mathcal{C}$. All data points are generated  from a \emph{ground-truth distribution} $p(\mv x, c)$ in an independent and identically distributed (i.i.d.) manner. We focus on \emph{pre-trained} SNN classification models, on which we make no assumptions in terms of accuracy or calibration. Furthermore, we assume the availability of a, typically  small, \emph{calibration data set} 
\begin{align}
    \mathcal{D}^{\rm cal}=\{\mv z[i]=(\mv x[i], c[i])\}_{i=1}^{|\mathcal{D}^{\rm cal}|}. \label{cal}
\end{align}
In practice, a new calibration data set may be produced periodically at test time to be reused across multiple test points $(\mv x, c)$   \cite{li2023unleashing, vovk2022algorithmic, angelopoulos2107gentle}.

\subsection{Taxonomy of SNN Classifiers}
\begin{figure*}[htp]
	\centering
	\includegraphics[width=5.2in]{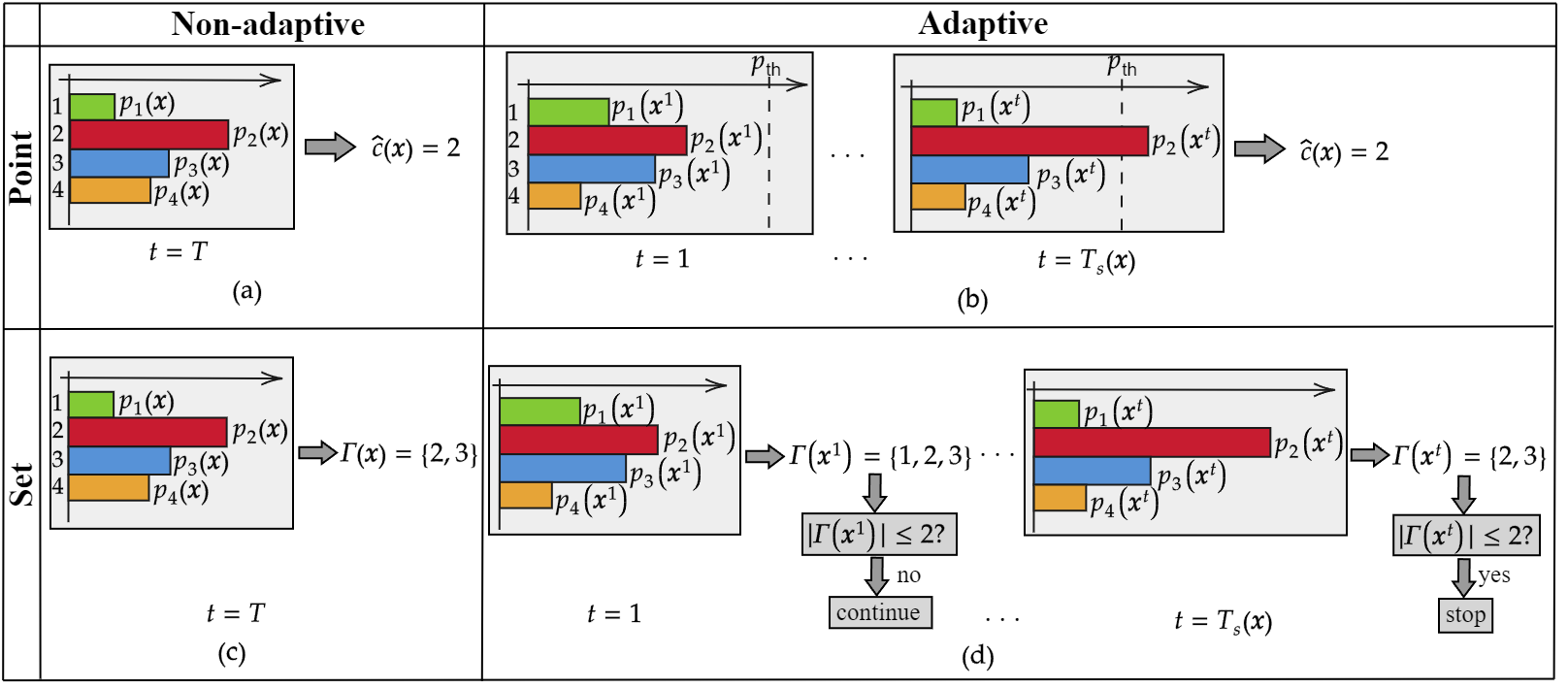}
	\caption{(a) A \emph{non-adaptive point} classifier   outputs a point decision $\hat{c}(\mv x)$ after having observed the entire time series $\mv x$. (b) An \emph{adaptive} point classifier  stops when the confidence level of the classifier passes a given threshold $p_{\text{th}}$, producing a classification decision  at an input-dependent time $T_s(\mv x)$. (c) A \emph{non-adaptive set} classifier produces a predicted set $\Gamma(\mv x)$ consisting of a subset of the class labels after having observed the entire time series $\mv x$. (d) The \emph{adaptive set} classifiers presented in this work stop at the earliest time $T_s(\mv x)$ when the predicted set $\Gamma(\mv x^{T_s(\scalebox{0.7}{\mv x})})$ is sufficiently informative, in the sense that its cardinality is below a given threshold $I_\text{th}$ (in the figure we set $I_\text{th}=2$). The proposed SpikeCP method can guarantee that the predicted set $\Gamma(\mv x)=\Gamma(\mv x^{T_s(\scalebox{0.7}{\mv x})})$ at the stopping time $T_s(\mv x)$ includes the true label with probability no smaller than the target probability $p_{\text{targ}}$. }
	\label{fig:adaptive_snns}
 \vspace{-0.5cm}
\end{figure*}

As detailed in Table~\ref{table:taxonomy} and Fig.~\ref{fig:adaptive_snns},  we distinguish SNN classifiers along two axes, namely adaptivity and decision type.

\begin{table}
\centering
\caption{Taxonomy of SNN classifiers}
\begin{tabular}{c||>{\centering\arraybackslash}p{2.4cm} |>{\centering\arraybackslash}p{2.7cm}}
\Xhline{2.2\arrayrulewidth}
\diagbox[width=10em, height=3em]{\textbf{decision type}}{\textbf{adaptivity}} & \textbf{non-adaptive} & \textbf{adaptive} \\
\hline \hline
\multirow{2.1}{*}{\textbf{point}} & ~~~conventional \newline (e.g., \cite{9317803}) & DC-SNN \cite{li2023unleashing}, \quad SEENN \cite{li2023seenn}  \\
\hline
\multirow{2}{*}{\textbf{set}} & ~~~SpikeCP\newline(this work) &  ~~~SpikeCP\newline(this work) \\
\Xhline{2.2\arrayrulewidth}
\end{tabular}
\label{table:taxonomy}
\vspace{-0.6cm} 
\end{table}

\emph{Adaptivity}: As shown in Fig.~\ref{fig:adaptive_snns}(a) and Fig.~\ref{fig:adaptive_snns}(c), a \emph{non-adaptive} classifier, having observed all the $T$ samples of the input sequence $\mv x$, makes a decision on the basis of the spike count vector $\mathbf{r}(\mv x^T) = \mathbf{r}(\mv x)$ or of the predictive probability vector $\mathbf{p}(\mv x^T)=\mathbf{p}(\mv x)$. In contrast, as seen in Fig.~\ref{fig:adaptive_snns}(b) and Fig.~\ref{fig:adaptive_snns}(d), an \emph{adaptive} classifier allows for the time $T_s(\mv x)$ at which a classification decision is produced, to be adapted to the difficulty of the input $\mv x$. For any given input $\mv x$, the \emph{stopping time} $T_s(\mv x)$ and the final decision produced at time $T_s(\mv x)$ depend on either the spike count vector $\mv r(\mv x^t)$ or on the predictive distribution vector $\mathbf{p}(\mv x^t)$ produced by the SNN classifier after having observed the first $t=T_s(\mv x)$ input samples $\mv x^t$ in \eqref{eq:firstsamples}. 


\emph{Decision type}: As illustrated in Fig.~\ref{fig:adaptive_snns}(a) and Fig.~\ref{fig:adaptive_snns}(b), for any given input $\mv x$, a conventional \emph{point classifier} produces as output a single estimate $\hat{c}(\mv x)$ of the label $c$ in a non-adaptive (Fig.~\ref{fig:adaptive_snns}(a)) or adaptive (Fig.~\ref{fig:adaptive_snns}(b)) way. In contrast, as seen in Fig.~\ref{fig:adaptive_snns}(c) and Fig.~\ref{fig:adaptive_snns}(d), a \emph{set classifier} outputs a decision in the form of  a \emph{subset} $\Gamma (\mv x)\subseteq \mathcal{C}$ of the $C$ classes  \cite{vovk2022algorithmic, angelopoulos2107gentle}, with the decision being non-adaptive (Fig.~\ref{fig:adaptive_snns}(c)) or adaptive (Fig.~\ref{fig:adaptive_snns}(d)). The \emph{predicted set} $\Gamma (\mv x)$ describes the classifier's estimate of the most likely candidate labels for input $\mv x$. Accordingly, a predicted set $\Gamma (\mv x)$ with a larger cardinality $|\Gamma (\mv x)|$ is less \emph{informative} than  one with a smaller (but non-zero)  cardinality. 


\subsection{Reliability, Latency, and Inference Energy}
\label{subsec:measures}
In this work, we study the performance of adaptive classifiers on the basis of the following metrics.

\emph{Reliability}: Given a \emph{target accuracy level} $p_{\rm targ}\in(0,1)$, an adaptive \emph{point} classifier is said to be \emph{reliable} if the accuracy of its decision is no smaller than the target  level $p_{\rm targ}$. This condition is stated as
\begin{align}
	&\mathrm{Pr}\big(c =\hat{c}(\mv x)\big) \geq p_{\rm targ},~ \nonumber\\\text{i.e.,}~ &\Delta R=  p_{\rm targ}-\mathrm{Pr}\big(c =\hat{c}(\mv x)\big) \leq 0,
    \label{reliap}
\end{align}
where  $\hat{c}(\mv x)$ is the decision made by the adaptive point classifier at time $T_s(\mv x)$ (see Fig. \ref{fig:adaptive_snns}(b)). In (\ref{reliap}), we have defined the \emph{reliability gap} $\Delta R$, which is positive for \emph{unreliable} classifiers and non-positive for \emph{reliable} ones (see Fig. \ref{fig:overall}(b)). In a similar manner, an adaptive \emph{set} predictor $\Gamma(\mv x)$ is reliable at the target accuracy level $p_{\rm targ}$ if the true class $c$ is included in the predicted set $\Gamma(\mv x)$, produced at the stopping time $T_s(\mv x)$, with probability no smaller than the desired accuracy level  $p_{\rm targ}$. This is written as
\begin{align}
	&\mathrm{Pr}\big(c \in \Gamma(\mv x )\big) \geq p_{\rm targ}, \nonumber\\\text{ i.e.}~&\Delta R=  p_{\rm targ}-\mathrm{Pr}\big(c =\Gamma(\mv x)\big) \leq 0,
    \label{relia}
\end{align}
where  $\Gamma(\mv x)$ is the decision made by the adaptive set classifier at time $T_s(\mv x)$ (see Fig. \ref{fig:adaptive_snns}(d)). The probabilities in \eqref{reliap} and \eqref{relia} are taken over the distribution of the test data point $(\mv x, c)$ and of the calibration data \eqref{cal}.


 \emph{Latency}: Latency is defined as  the average stopping time $\mathbb{E}[T_s(\mv x)]$, where the expectation is taken over the same distribution as for \eqref{reliap} and \eqref{relia}. 

\emph{Inference energy}: As a proxy for the energy consumption of the SNN classifier at inference time, we follow the standard approach also adopted in, e.g.,  \cite{10016643, 9997098}, of counting the average number of spikes, denoted as $\mathbb{E}[S(\mv x)]$, that are produced internally by the SNN classifier prior to producing a decision.

\subsection{Spiking Neural Network Model} \label{snn model}
In this work, we adopt the standard LIF  neural model known as  \emph{spike response model} (SRM) \cite{skatchkovsky2021spiking}. LIF model is the most commonly adopted neural model for SNNs, simulating the behavior of biological neurons that integrate stimuli over time and fire a spike once a certain threshold on the integral is reached. Consider a set of  spiking neurons indexed via integers in  set $\mathcal{K}$. Each spiking neuron $k \in \mathcal{K}$ outputs a binary signal $b_{k,t}\in\{0,1\}$ at time $t=1,..., T$, with $b_{k,t}=1$ representing the firing of the spike and $b_{k,t}=0$ an idle neuron at time $t$.  It receives inputs from a subset of neurons $ \mathcal{N}_k$ through directed links, known as \emph{synapses}. Accordingly, neurons in set $\mathcal{N}_k$ are referred to as \emph{pre-synaptic} with respect to neuron $k$; while neuron $k$ is said to be  \emph{post-synaptic} for any neuron $j \in \mathcal{N}_k$. For a fully-connected layered SNN, as assumed in the experiments of this paper, the set of pre-synaptic neurons, $\mathcal{N}_k$, for a neuron $k$ in a given layer consists of the entire  set of indices of the neurons in the previous layer. 
 
Following the SRM, each neuron $k$ maintains an internal analog state variable $o_{k,t}$, known as the \emph{membrane potential}, over time $t$. The membrane potential $o_{k,t}$ evolves as the sum of the responses of the synapses to the incoming spikes produced by the pre-synaptic neurons, as well as of the response of the neuron itself to the spikes it produces. Mathematically, the evolution of the membrane potential is given as 
\begin{align}	o_{k,t}=\sum_{j\in\mathcal{N}_k}w_{k,j}\cdot(\alpha_t * b_{j,t})+\beta_t * b_{k,t},
	\label{potential}
\end{align}
where $w_{k,j}$ is a learnable synaptic weight between neuron $j\in\mathcal{N}_k$ and neuron $k$; $\alpha_t$ represents a filter applied to the spiking signals produced by each pre-synaptic neurons; $\beta_t$ is the filter applied to its own spiking output; and ``$*$'' denotes the convolution operator. 

Typical choices for synaptic filters include the first-order feedback filter $\beta_t=\exp(-t/\tau_{\rm ref})$, and the second-order synaptic filter $\alpha_t=\exp(-t/{\tau_{\rm mem}})-\exp(-t/{\tau_{\rm syn}})$, for $t=1,2,...$, with finite positive constants $\tau_{\rm ref}$, $\tau_{\rm mem}$, and $\tau_{\rm syn}$ \cite{neftci2019surrogate}. Each neuron $k$ outputs a spike at time step $t$ whenever its membrane potential crosses a fixed threshold $\vartheta$, i.e.,
 \begin{align}
	b_{k,t}=\Theta(o_{k,t}-\vartheta),
	\label{spike}
\end{align}
where $\Theta(\cdot)$ is the Heaviside step function. 

The synaptic weights $w_{k,j}$ in \eqref{potential} between any neurons $k \in \mathcal{K}$ and the corresponding pre-synaptic neurons $j \in \mathcal{N}_k$ constitute the model parameters to be optimized during training. Accordingly, we write as $\mv \theta = \{\{w_{k,j}\}_{ j\in\mathcal{N}_k}\}_{k\in\mathcal{K}}$ the vector of model parameters of the SNN.

\section{Adaptive Point Classification}
\label{sec:adaptive_point_classification}
 In this section, we review, for reference, the adaptive point classifiers introduced in \cite{li2023unleashing} and \cite{li2023seenn}, which are referred to as \emph{dynamic-confidence SNN} (DC-SNN) and \emph{stopping-policy SNN} (SP-SNN), respectively.

\emph{DC-SNN} \citep{li2023unleashing}: As illustrated in Fig.~\ref{fig:adaptive_snns}(b), DC-SNN produces a decision at the first time $t$ for which the maximum confidence level across all possible classes is larger than a fixed \emph{target confidence level} $p_{\textrm{th}}\in(0,1)$. Accordingly, the stopping time is given by 
\begin{align}
T_s(\mv x)=&\min_{t \in \{1,...,T\}} t  ~\text{ s.t.}~~ \max_{c\in\mathcal{C}} p_{c}(\mv x^t)\geq p_{\textrm{th}}, \label{stop}   
\end{align}  
if there is a time $t<T$ that satisfies the constraint; and $T_s(\mv x)=T$ otherwise. The rationale for this approach is that, by (\ref{stop}), if $T_s(\mv x)<T$, the classifier has a confidence level no smaller than $p_{\textrm{th}}$ on the decision 
\begin{equation}
    \hat{c}(\mv x)=\arg\max_{c\in\mathcal{C}} p_{c}(\mv x^{T_s(\scalebox{0.7}{\mv x})}). \label{decision}
\end{equation} 
If the SNN classifier is \emph{well calibrated}, the confidence level coincides with the true accuracy of the decision given by the class $\argmax_{c\in \mathcal{C}} p_c (\mv x^t)$ at all times $t$. Therefore, setting the target confidence level $p_{\textrm{th}}$ to be equal to the target accuracy $p_\text{targ}$, i.e., $p_\text{th} = p_\text{targ}$, guarantees a zero, or negative, reliability gap for the adaptive decision (\ref{decision}) when $T_s(\mv x)<T$. However, as discussed in Sec. \ref{Sec:intro}, the assumption of calibration is typically not valid (see Fig. \ref{fig:overall}(b)).  To address this problem, reference  \cite{li2023unleashing} introduced a solution based on the use of a calibration data set.

 Specifically, DC-SNN evaluates the empirical accuracy of the decision (\ref{decision}), i.e., $\hat{\mathcal{A}}^{\textrm{cal}}(p_\mathrm{th})=|\mathcal{D}^{\rm cal}|^{-1}\sum_{i=1}^{|\mathcal{D}^{\rm cal}|}\mathbbm{1}(\hat{c}(\mv x[i])=c[i])$, where $\mathbbm{1}(\cdot)$ is the indicator function,  for a grid of possible values of  the target confidence level $p_{\textrm{th}}$. Then, it  chooses the minimum value $p_{\textrm{th}}$ that ensures the inequality $\hat{\mathcal{A}}^{\textrm{cal}}(p_\mathrm{th})\geq p_\text{targ}$, so that the calibration accuracy exceeds the target accuracy level $p_\text{targ}$; or the smallest value $p_{\rm th}$ that maximizes $\hat{\mathcal{A}}^{\textrm{cal}}(p_{\rm th})$ if the constraint  $\hat{\mathcal{A}}^{\textrm{cal}}(p_\mathrm{th})\geq p_\text{targ}$ cannot be met.

\emph{SP-SNN} \citep{li2023seenn}: SP-SNN defines a parameterized \emph{policy} $\mv \pi(\mv x|\mv \phi)$, implemented using a separate artificial neural network (ANN),  that maps the input sequence $\mv x$  to a probability distribution $\mv \pi(\mv x|\mv \phi) = [\pi_1(\mv x|\mv \phi), ..., \pi_T(\mv x|\mv \phi)]$ over the $T$ time steps, where $\mv \phi$ is the trainable parameter vector of the ANN. Accordingly, given input $\mv x$, the stopping time is drawn using the policy $\mv \pi(\mv x|\mv \phi)$ as $T_s(\mv x) \sim \mv \pi(\mv x|\mv \phi)$. 

Unlike DC-SNN, which uses a pre-trained SNN, the policy in SP-SNN is optimized jointly with the SNN based on an available training data set
\begin{align}
    \mathcal{D}^{\rm tr}=\{(\mv x^{\rm tr}[i], c^{\rm tr}[i])\}_{i=1}^{|\mathcal{D}^{\rm tr}|} \label{trainingdata}
\end{align}
of $|\mathcal{D}^{\rm tr}|$ examples, 
whose data points are i.i.d. as for the calibration data set \eqref{cal} and for the test data. Furthermore, unlike DC-SNN, SP-SNN does not make use of calibration data. 

Optimization in SP-SNN targets an objective function that depends on a combination of latency and accuracy. To be specific, given a training example $(\mv x, c) \in \mathcal{D}^\text{tr}$, SP-SNN takes an action $T_s(\mv x)$ derived by the policy, from which a \emph{reward} 
\begin{align}\label{eq:reward}
R\big(T_s(\mv x)\big)=\begin{cases}
1/2^{T_s(\scalebox{0.7}{\mv x})}, & \mbox{$\hat{c}(\mv x)=c$}, \\ 
-\zeta, & \mbox{otherwise},
\end{cases}
\end{align} 
is provided to SP-SNN to optimize the policy ANN, where $\zeta$ is a positive constant. 
Accordingly, if the prediction is correct, i.e., if $\hat{c}(\mv x)=c$, the reward (\ref{eq:reward}) favors lower latencies by assigning a larger reward to a policy that produces a decision at an earlier time $T_s(\mv x)$. Conversely, if the prediction is wrong, a penalty $\zeta$ is applied. 

Accuracy in SP-SNN is accounted for via the standard \emph{cross-entropy loss}. For an example $(\mv x, c)$ at stopping time $T_s(\mv x)$, this is defined  as 
\begin{align}
    L(\mv x^{T_s(\scalebox{0.7}{\mv x})})=-\log p_c(\mv x^{T_s(\scalebox{0.7}{\mv x})}),
\end{align} 
where probability $p_c(\mv x^{T_s(\scalebox{0.7}{\mv x})})$ is defined in \eqref{prob}. Accordingly, SP-SNN jointly optimizes the SNN parameters $\mv \theta$ (see Sec. \ref{exp}) and the policy network parameters $\mv \phi$ by addressing the problem
\begin{align}
\min_{\scalebox{0.7}{\mv \theta}, \scalebox{0.7}{\mv \phi}} \sum_{(\scalebox{0.7}{\mv x},c)\in\mathcal{D}^{\rm tr}}\mathbb{E}[-R\big(T_s(\mv x)\big)+L(\mv x^{ T_s(\scalebox{0.7}{\mv x})})], \label{joint_op}   
\end{align}  
where the expectation is taken over the probability distribution $\mv \pi(\mv x|\mv \phi)$. The problem is tackled via an alternate application of reinforcement learning for the optimization of parameters $\mv \phi$ and of supervised learning for the optimization of parameters $\mv \theta$.




\section{SpikeCP: Reliable Adaptive Set Classification} \label{spikecp}

The adaptive point classifiers reviewed in the previous section are generally characterized by a positive reliability gap (see Fig. \ref{fig:overall}(a)), unless the underlying SNN classifier is well calibrated or unless the calibration data set is large enough to ensure a reliable estimate of the true accuracy. In this section, we introduce \emph{SpikeCP}, a novel inference methodology for adaptive classification that wraps around any pre-trained SNN model, guaranteeing the reliability requirement \eqref{relia} -- and hence a zero, or negative, reliability gap -- irrespective of the quality of the SNN classifier and of the amount of calibration data. In the next section we discuss how to potentially improve the performance of SpikeCP by training tailored SNN models.


\subsection{Stopping Time}
\label{sec:spikeCP}
SpikeCP pre-determines a subset of possible stopping times, referred to as \emph{checkpoints}, in set $\mathcal{T}_s \subseteq \{1,...,T\}$. Set $\mathcal{T}_s \subseteq \{1,...,T\}$ always includes the last time $T$, and adaptivity is only possible if the cardinality of set $\mathcal{T}_s$ is strictly larger than one. At each time $t\in\mathcal{T}_s$, using the local spike count variables $\mathbf{r}(\mv x^t)$ or the global predictive probabilities $\mathbf{p}(\mv x^t)$, SpikeCP produces a candidate predicted set $\Gamma(\mv x^t)\subseteq \mathcal{C}$. Then, as illustrated in Fig. \ref{fig:adaptive_snns}(d), the cardinality $|\Gamma(\mv x^t)|$ of the candidate predicted set $\Gamma(\mv x^t)$ is compared with a threshold $I_{\rm th}$. If we have the inequality  \begin{equation}
\label{eq:thresholdset}
|\Gamma(\mv x^t)|\leq I_{\rm th},
\end{equation}
the predicted set is deemed to be  sufficiently \emph{informative}, and SpikeCP stops processing the input to produce set $\Gamma(\mv x^t)$ as the final decision $\Gamma(\mv x)$. As we detail next and as illustrated in Fig. \ref{fig:overall}(c), the candidate predicted sets $\Gamma(\mv x^t)$ are constructed in such a way to ensure a non-positive reliability gap simultaneously for \emph{all} checkpoints, and hence also at the stopping time. The overall procedure of SpikeCP is summarized in Algorithm 1.

To construct the candidate predicted set $\Gamma(\mv x^t)$ at a checkpoint $t\in\mathcal{T}_s$, SpikeCP follows the \emph{split}, or \emph{validation-based}, CP procedure proposed in  \cite{vovk2022algorithmic} and reviewed in \cite{angelopoulos2107gentle, cp2023}. Accordingly, using the local spike counts $\mathbf{r}(\mv x^t)$ or the global probabilities $\mathbf{p}(\mv x^t)$,  SpikeCP produces a so-called \emph{non-conformity (NC) score} vector $\mathbf{s}(\mv x^t)=[s_1(\mv x^t),...,s_C(\mv x^t)]$. Each entry $s_c(\mv x^t)$ of this vector is a measure of the \emph{lack of confidence} of the SNN classifier  in label $c$ given input $\mv x^t$. The candidate predicted set $\Gamma(\mv x^t)$ is then obtained by including all labels $c\in\mathcal{C}$ whose NC score $s_c(\mv x^t)$ is no larger than a threshold $s_{\rm th}^t$, i.e., 
\begin{align}
     \Gamma(\mv x^t)=\{c\in \mathcal{C}: s_c(\mv x^t)\leq s^t_{\rm th} \}.
     \label{cpsett}
 \end{align}
As described in Sec. \ref{eth}, the threshold $s^t_{\rm th}$ is evaluated as a function of the target accuracy level $p_{\rm targ}$, of the the calibration set $\mathcal{D}^\text{cal}$, and of the number of checkpoints $|\mathcal{T}_s|$.

 We consider two NC scores, one locally computable at the output neurons and one requiring coordination among the output neurons.  The \emph{local NC score} is defined as \begin{align}
     s_c(\mv x^t)=t - r_{c}(\mv x^t).
 	\label{ncs}
 \end{align} 
 Intuitively, class $c$ is assigned a lower NC score \eqref{ncs} -- and hence a higher degree of confidence -- if the spike count variable $r_{c}(\mv x^t)$ is larger. In contrast, the \emph{global NC score} is given by the standard \emph{log-loss}
 \begin{align}
     s_c(\mv x^t)=-\log p_{c}(\mv x^t).
 	\label{nc}
\end{align}

\subsection{Evaluation of the Threshold } \label{eth}

\begin{figure}
\centering
\includegraphics[width=3.2in]{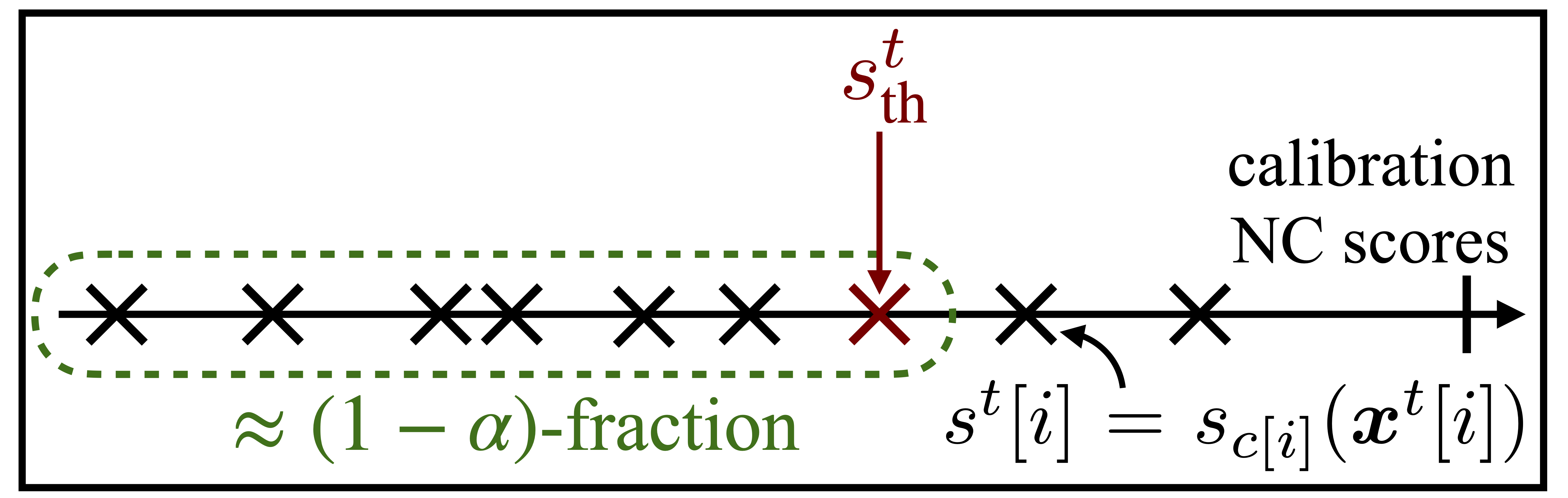}
\caption{ CP meets condition \eqref{eq:percp} by choosing the threshold $s_{\rm th}^t$ in \eqref{cpsett} as the $\lceil (1-\alpha)(|\mathcal{D}^\text{cal}|+1)\rceil$-th smallest value among the NC scores evaluated in the calibration set.} 
\label{fig:soft_quan_diagram}
\vspace{-0.5cm} 
\end{figure}

As we detail in this subsection, the threshold $s^t_{\rm th}$ in \eqref{cpsett} is evaluated based on the calibration data set $\mathcal{D}^{\rm cal}$  with the goal of ensuring the reliability condition (\ref{relia}) for a target accuracy level $p_{\textrm{targ}}$. The general methodology follows CP, with the important caveat that, in order to ensure a non-positive reliability gap simultaneously at all checkpoints, a form of \emph{Bonferroni correction} is applied. Alternative, heuristic, corrections are also described at the end of this section.

Let us define as $1-\alpha$, with $\alpha\in (0,1)$, an auxiliary \emph{per-checkpoint accuracy level}. Suppose that we can guarantee the \emph{per-checkpoint reliability condition}  
\begin{equation}
\label{eq:percp}
\mathrm{Pr}(c \in \Gamma(\mv x^{t} )) \geq 1-\alpha
\end{equation} 
for all checkpoints $t\in \mathcal{T}_s$. In (\ref{eq:percp}), the probability is taken over the distribution of the test and calibration data. We will see below that this condition can be guaranteed by leveraging the toolbox of CP. Then by De Morgan's law and the union bound, we also have the reliability condition 
\begin{equation}
\label{eq:simult}\mathrm{Pr}(c \in \Gamma(\mv x^{t} ) \textrm{ for all }t\in\mathcal{T}_s) \geq 1-|\mathcal{T}_s|\alpha, 
\end{equation} 
which applies simultaneously across all checkpoints. This inequality implies that we can guarantee the condition (\ref{relia}) by setting $\alpha=(1-p_{\textrm{targ}})/|\mathcal{T}_s|$, since the stopping point $T_s(\mv x)$ is in set $\mathcal{T}_s$ by construction. This is a form of \emph{Bonferroni correction}, whereby the target accuracy for the test carried out at each checkpoint is increased in order to ensure reliability simultaneously for the tests at all checkpoints \cite{vovk2021values}. This increase is linear in the number of checkpoints $|\mathcal{T}_s|$, and it guarantees the desired reliability condition irrespective of the underlying distribution of the data, as long as the per-checkpoint inequality (\ref{eq:percp}) is satisfied.

The remaining open question is how to ensure the per-checkpoint reliability condition (\ref{eq:percp}). To address this goal, we  follow the standard CP procedure. Accordingly, during an \emph{offline} phase, for each calibration data point $(\mv x[i], c[i])$, with $i=1,..., |\mathcal{D}^{\rm cal}|$, SpikeCP computes  the NC score $s^t[i]=s_{c[i]}(\mv x^t[i])$ at each checkpoint  $t\in\mathcal{T}_s$. The calibration NC scores $\{s^t[i]\}_{i=1}^{|\mathcal{D}^{\rm cal}|}$ are ordered from smallest to largest, with ties broken arbitrarily, separately for each checkpoint $t$. Finally, the threshold $s_{\rm th}^t$ is selected to be approximately equal to  the smallest value that is larger than a fraction $(1-\alpha)$ of the calibration NC scores (see Fig.~3). 
More precisely, assuming $\alpha\geq 1/(|\mathcal{D}^\text{cal}|+1)$ we set \cite{vovk2022algorithmic, angelopoulos2107gentle}
\begin{align}
    \label{eq:s_th}
    s_{\rm th}^t = \lceil (1-\alpha)(|\mathcal{D}^\text{cal}|+1)\rceil&\text{-th smallest value} \notag \\
    &~ \text{in the set } \{ s^t[i] \}_{i=1}^{|\mathcal{D}^\text{cal}|},
\end{align}
while for $\alpha < 1/(|\mathcal{D}^\text{cal}|+1)$ we set $s_\text{th}^t=\infty$. This is illustrated in Fig. \ref{fig:soft_quan_diagram}. 

\setlength{\textfloatsep}{0pt}
\begin{algorithm}[t!]
	\caption{SpikeCP}\label{Al 2}	
	\SetKwInOut{Input}{Input}
	\Input{Pre-trained SNN classifier;  calibration set $\mathcal{D}^{\rm cal}$; checkpoint candidates $\mathcal{T}_s$; target accuracy level $p_{\mathrm{targ}} \in (0,1)$; target set size (informativeness) $I_\text{th}$; and test input $\mv x$}
	\SetKwInOut{Output}{Output}
	\Output{Adaptive set classification  $\Gamma(\mv x)$ at time $T_s(\mv x)$ satisfying the reliability condition (\ref{relia})}
 \underline{\emph{Offline phase}}:\\
    Compute the NC scores $s^t[i]$ for all calibration data points $i=1,...,|\mathcal{D}_{\rm cal}|$ in set $\mathcal{D}^{\rm cal}$ and for all checkpoints  $t\in\mathcal{T}_s$ based on \eqref{ncs} or \eqref{nc} \\
    For each checkpoint $t \in \mathcal{T}_s$, obtain the threshold $s^t_{\rm th}$ as the $\lceil (1-\alpha)(|\mathcal{D}^\text{cal}|+1)\rceil$-th smallest NC score in the set $\{s^t[i]\}_{i=1}^{|\mathcal{D}^{\rm cal}|}$ with $\alpha = (1-p_\text{targ})/|\mathcal{T}_s|$ if $\alpha \geq 1/(|\mathcal{D}^\text{cal}|+1)$; otherwise set $s_\text{th}^t=\infty$\\
    \underline{\emph{Test time}}: \\
    \For{each checkpoint time $t\in\mathcal{T}_s$}{
        Generate the set predictor $\Gamma(\mv x^t)$ based on \eqref{cpsett} with threshold $s_{\rm th}^t$ \\
        \If{$|\Gamma(\mv x^t)| \leq I_\text{th}$}{        
        Exit}
    } 
    Set $\Gamma(\mv x)=\Gamma(\mv x^t)$ and stopping time $T_s(\mv x)=t$  \\
    \textbf{Return:} $\Gamma(\mv x)$
\end{algorithm}

\subsection{Reliability Guarantees of SpikeCP}

In this subsection, we show that SpikeCP, as summarized in Algorithm 1, satisfies the reliability condition (\ref{relia}).

\begin{theorem}[Reliability of SpikeCP]
\label{theor}
The adaptive decision $\Gamma(\mv x)=\Gamma(\mv x^{T_s(\scalebox{0.7}{\mv x})})$  produced by SpikeCP, as described in Algorithm 1, satisfies the reliability condition (\ref{relia}), and hence has a non-positive reliability gap, i.e., $\triangle R \leq 0$. 
\end{theorem}

\begin{proof} 
By the properties of CP, the threshold \eqref{eq:s_th} ensures the per-checkpoint reliability condition (\ref{eq:percp}) (see, e.g., \citep[Theorem 1]{tibshirani2019conformal} and \cite{vovk2022algorithmic, kuchibhotla2020exchangeability, lei2018distribution}). By applying De Morgan's law and the union bound, we ensure that the reliability condition \eqref{eq:simult} holds for all checkpoints. Consequently, we can conclude that the reliability condition (\ref{relia}) is met by setting $\alpha=(1-p_{\textrm{targ}})/|\mathcal{T}_s|$. We refer to the Appendix for further details. 
\end{proof}

\subsection{An Alternative Heuristic Threshold Selection} \label{sec:simes}
The theoretical guarantees of SpikeCP in Theorem~\ref{theor}  rely on the Bonferroni correction that sets the  per-checkpoint target accuracy level to $1-\alpha=1-(1-p_{\textrm{targ}})/|\mathcal{T}_s|$. This requirement  becomes increasingly stricter, and hence harder to satisfy, as the number of checkpoints $|\mathcal{T}_s|$ increases. However, having  a large number of checkpoints may be advantageous by enhancing the granularity of delay adaptivity. 


In this subsection, we introduce an alternative, heuristic, choice for the per-checkpoint reliability condition based on Simes correction  \cite{rodland2006simes}. The approach sets a different target $ 1-\alpha_t$ for each checkpoint $t\in\mathcal{T}_s$, by imposing the constraint  \begin{equation}
\label{eq:percp_simes}
\mathrm{Pr}(c \in \Gamma(\mv x^{t} )) \geq 1-\alpha_t
\end{equation} 
in lieu of the constant-target condition (\ref{eq:percp}). For each time step $t \in \mathcal{T}_s$, let us define $i_t$ for the index that runs across the checkpoints as $i_t = \sum_{t' \in \mathcal{T}_s} \mathbbm{1}(t' \leq t)$. Then, the target reliability for the checkpoint at time $t\in \mathcal{T}_s$ is set to  $1-\alpha_t$ with
\begin{align}
    \label{eq:simes_per_checkpoint_acc}
    \alpha_t = i_t \cdot \frac{(1-p_{\rm targ})}{|\mathcal{T}_s|}.
\end{align}
Accordingly, for the first checkpoint $t$, with $i_1=1$, the  target coincides with that obtained from Bonferroni correction, i.e., $\alpha_{t} = (1-p_\text{targ})/|\mathcal{T}_s|$; while for the last checkpoint, with $i_t=|\mathcal{T}_s|$, it corresponds to the target accuracy level, i.e., $\alpha_{t} = 1-p_\text{targ}$.





Using Simes correction (\ref{eq:simes_per_checkpoint_acc}) in step 3 in Algorithm 1 in lieu of $\alpha_t = (1-p_\text{targ})/|\mathcal{T}_s|$,  yields an alternative version of SpikeCP that is guaranteed to meet the reliability condition  \eqref{reliap} only under additional assumptions that are hard to verify in practice (see Appendix). One of such assumptions is that the accuracy of SNN never decreases with increased time steps, as posited, e.g., in  \cite[Assumption 3.1]{li2023seenn}. Given this limitation, we propose Simes correction here merely as a heuristic, which may yield some practical gains as demonstrated in Sec. \ref{exp:bs} (see Fig. \ref{r7}).

\section{SpikeCP-Based Training} \label{spikecp+}
While SpikeCP provides guarantees on the reliability of its set-valued decisions  irrespective of the quality of the pre-trained SNN (see Theorem~\ref{theor}), the achievable trade-offs between average delay and energy consumption, on the one hand, and informativeness of the set predictor, on the other, generally depend on the performance of the underlying SNN-based classifier. In this section, we introduce a training strategy -- referred to as \emph{SpikeCP-based training} -- that, unlike conventional learning algorithms for SNNs (see, e.g., \cite{wade2010swat,skatchkovsky2021spiking}), targets directly the performance of the SNN when used in conjunction with SpikeCP. 

\subsection{Training Objective}

In order to describe the training objective of SpikeCP-based training, we start by recalling from Sec. \ref{sec:spikeCP} that the stopping time of SpikeCP is determined by the size $|\Gamma(\mv x^t)|$ of the predicted set $\Gamma(\mv x^t)$ for input $\mv x$ as per the threshold rule (\ref{eq:thresholdset}) with target set size $I_{\rm th}$. Therefore,  to reduce the average latency, one can train the SNN with the aim at minimizing the sizes $|\Gamma(\mv x^t)|$ of the predicted sets $\Gamma(\mv x^t)$ in (\ref{cpsett}) produced by SpikeCP over time instants $t$ with the set $\mathcal{T}_s$ of candidates checkpoints.

To this end, the model parameters $\mv \theta$ are optimized on the basis of the training set \eqref{trainingdata}.
Specifically, in order to mimic the test-time distinction between calibration and test data leveraged by SpikeCP, we randomly partition the training set $\mathcal{D}^{\rm tr}$ into two disjoint subsets $\mathcal{D}^\text{tr,cal}$ and $\mathcal{D}^\text{tr,te}$  with $\mathcal{D}^\text{tr,cal} \cap \mathcal{D}^\text{tr,te}= \emptyset$ and $\mathcal{D}^\text{tr,cal} \cup \mathcal{D}^\text{tr,te}=\mathcal{D}^{\rm tr}$. 

Given a data set split $(\mathcal{D}^\text{tr,cal}, \mathcal{D}^\text{tr,te})$, we run SpikeCP (Algorithm 1) with $\mathcal{D}^\text{tr,cal}$ in lieu of the calibration data  $\mathcal{D}^\text{cal}$, and with the input parts of the  data points in the set $\mathcal{D}^\text{tr,te}$ as the test inputs $\mv x$. For each such test input $\mv x$ in $\mathcal{D}^\text{tr,te}$, SpikeCP returns the predictive set ${\Gamma}(\mv x^{t})$ for all checkpoints $t\in\mathcal{T}_s$. In line with the motivation explained in the previous paragraph, we consider the set sizes $|{\Gamma}(\mv x^{t})|$ for all time instants $t$ in the  checkpoint set $\mathcal{T}_s$ as the target of the training process. 

To quantify the mentioned predictive set sizes using training data, we define the \emph{efficiency training loss}
\begin{equation}\label{eq:efficiency_loss} 
\mathcal{L}^{E}(\mv \theta)=\sum_{\substack{\mathcal{D}^\text{tr,cal} \subset \mathcal{D}^\text{tr}\\ \mathcal{D}^\text{tr,te} = \mathcal{D}^\text{tr} \setminus \mathcal{D}^\text{tr,cal}}} \sum_{{\scriptsize (\mv x,c)} \in \mathcal{D}^\text{tr,te}}\sum_{t\in \mathcal{T}_s}|\Gamma(\mv x^t)|.  
\end{equation} 
 The outer sum in \eqref{eq:efficiency_loss} is over a number of splits realized by randomly sampling the subset   $\mathcal{D}^\text{tr,cal} \subset \mathcal{D}^\text{tr}$ for a fixed given number of calibration data points $|\mathcal{D}^\text{tr,cal}|<|\mathcal{D}^\text{tr}|$; the middle sum is over the test data points in set $\mathcal{D}^\text{tr,te} = \mathcal{D}^\text{tr} \setminus \mathcal{D}^\text{tr,cal}$; and  the inner sum is over the time instants in the checkpoint set $\mathcal{T}_s$.  

The efficiency training loss $\mathcal{L}^{E}(\mv \theta)$ in \eqref{eq:efficiency_loss} does not make use of the labels of the test data sets, and it does not directly target the accuracy of the SNN classifier. In a manner somewhat similar to the criterion \eqref{joint_op} used by SP-SNN, we hence propose to  complement the efficiency training loss with the standard \emph{cross-entropy training  loss} as
\begin{equation}
\label{eq:ce_loss}\mathcal{L}^{C}(\mv \theta)=-\sum_{\substack{\mathcal{D}^\text{tr,cal} \subset \mathcal{D}^\text{tr}\\ \mathcal{D}^\text{tr,te} = \mathcal{D}^\text{tr} \setminus \mathcal{D}^\text{tr,cal}}} \sum_{{\scriptsize (\mv x, c)} \in \mathcal{D}^\text{tr,te}}\sum_{t\in \mathcal{T}_s}\log p_c(\mv x^t),
\end{equation} 
where $p_c(\mv x^t)$ is the probability value assigned by the model to input $\mv x^t$ for class $c$ using \eqref{prob}. The sums in \eqref{eq:ce_loss} are evaluated as for the efficiency training  loss (\ref{eq:efficiency_loss}).



Overall, we propose to optimize the parameter vector $\mv \theta$ of SNN by addressing the problem
\begin{align}
    \min_{\scriptsize \mv \theta} \mathcal{L}^{C}(\mv \theta) + \lambda \mathcal{L}^{E}(\mv \theta)\label{oloss}
\end{align}
for a hyperparameter $\lambda \geq 0$ that dictates the trade-off between cross-entropy and efficiency criteria.  With  $\lambda=0$ and $\mathcal{T}_s=\{T\}$, this training objective recovers the conventional cross-entropy evaluated at the last time instant adopted in most of the literature on SNN-based  classification (see, e.g.,  \cite{10016643,wade2010swat}).





\subsection{Training}
\label{subsec:spikecp_based_tr_B}

The gradient of the standard cross-entropy objective $\mathcal{L}^{C}(\mv \theta)$ can be approximated via well-established surrogate gradient methods that apply the straight-through estimator of the gradient \cite{eshraghian2021training, skatchkovsky2021spiking}. Accordingly, when applying backpropagation, while the forward pass uses the actual non-differentiable activation model \eqref{spike} of the SRM neurons, the backward pass replaces the non-differentiable spiking threshold function \eqref{spike} with a smooth sigmoidal function \cite{eshraghian2021training, skatchkovsky2021spiking}. For each neuron $k$ at time $t$, this yields the differentiable activation\begin{align}
	\hat{b}_{k,t}=\sigma(o_{k,t}-\vartheta),
	\label{sspike}
\end{align}
where the Heaviside step function $\Theta(\cdot)$ in \eqref{spike} is replaced by  the sigmoid function $\sigma(x)=1/(1+e^{-x})$. 

Given the availability of surrogate gradient methods, the main new challenge in tackling problem (\ref{oloss}) lies in the evaluation of the gradient  of the criterion $\mathcal{L}^E(\mv \theta)$. The rest of this section focuses on this problem.

The efficiency training loss $\mathcal{L}^E(\mv \theta)$  in \eqref{eq:efficiency_loss} depends on the cardinality $|\Gamma(\mv x^t)|$, which  is a non-differentiable function of the model parameters $\mv \theta$, even when considering the surrogate SNN model with activation function in \eqref{sspike}.  In fact, the NC scores $s_c(\mv x^t)$ in \eqref{ncs} or \eqref{nc} are differentiable in $\mv \theta$ under the surrogate model \eqref{sspike}, but this is not the case for the cardinality  $|\Gamma(\mv x^t)|$ of the predicted set. 

To see this, observe that cardinality $|\Gamma(\mv x^t)|$ is obtained via a cascade of two non-differentiable functions of the scores $s^t[i]$, $i=1,..., |\mathcal{D}^{\rm cal}|$: (\emph{i}) \emph{Sorting}: By Algorithm 1, SpikeCP sorts the calibration scores $s^t[i]$, $i=1,..., |\mathcal{D}^{\rm cal}|$, to obtain the threshold $s_\text{th}^t$ via \eqref{eq:s_th} at each checkpoint time $t\in\mathcal{T}_s$;  (\emph{ii})
 \emph{Counting}: The cardinality $|\Gamma(\mv x^t)|$ of the set predictor is obtained by counting the number of labels $c$ whose score $s_c(\mv x^{t})$ is no larger than the threshold $s^t_{\rm th}$, i.e., $|\Gamma(\mv x^t)|= \sum_{c=1}^C\mathbbm{1} (s_c(\mv x^{t}) \leq  s_\text{th}^t  )$.

In the next subsection, we introduce a differentiable approximation $|\hat{\Gamma}(\mv x^t)|$ of the cardinality function $|\Gamma(\mv x^t)|$ under the smooth activation \eqref{asize}. The approach follows prior art on  CP-aware training \cite{stutz2021learning, einbinder2022training, park2022few}.

\subsection{Differentiable Threshold and Set Cardinality }
\label{subsec:diff_th}
The threshold $s_\text{th}^t$ in \eqref{eq:s_th} amounts to the $(1-\alpha)$-empirical quantile of the calibration scores $s^t[i]$, $i=1,..., |\mathcal{D}^{\rm cal}|$. Given $\mathcal{D}^\text{tr,cal}$, this can be obtained as the solution of the problem
\begin{align}
    \label{eq:sol_pinball}
    s_\text{th}^t = \arg\min_{s \in  \{s^t[i]\}_{i=1}^{|\mathcal{D}^\text{tr,cal}|}  } \big( \rho_{1-\alpha}(s|\{s^t[i]\}_{i=1}^{|\mathcal{D}^\text{tr,cal}|}\cup \{\infty\})\big)
\end{align}
where we have defined the \emph{pinball loss} as
\begin{align}
   \rho_{1-\alpha}(a|\{a[i]\}_{i=1}^{M})= &~ \alpha\sum_{i=1}^{M}\text{ReLU}(a-a[i])+ \notag
   \\&~ (1-\alpha)\sum_{i=1}^{M}\text{ReLU}(a[i]-a),
	\label{pin}
\end{align}
for $M$ real numbers $\{a[i]\}_{i=1}^M$ with $\text{ReLU}(a)=\max(0,a)$.

The solution of problem \eqref{eq:sol_pinball} can be approximated by replacing the minimum with a \emph{soft minimum} function $\delta(x_i)=e^{-x_i}/\sum_j e^{-x_j}$. Accordingly, a differentiable estimate of the threshold $s^t_{\rm th}$ can be written as \cite{park2022few}
\begin{align}
    \hat{s}_\text{th}^t = \sum_{i=1}^{|\mathcal{D}^{\rm tr,cal}|+1}s^t[i] \delta\big(\frac{\rho_{1-\alpha}(s^t[i]|\{s^t[i]\}_{i=1}^{|\mathcal{D}^{\rm tr,cal}|+1})}{c_Q}\big),
    \label{eq:smooth_min}
\end{align}
where we have defined $s^t[{|\mathcal{D}^{\rm tr, cal}|+1}]=\max(\{s^t[i]\}_{i=1}^{|\mathcal{D}^{\rm tr,cal}|})+\beta$ for some sufficiently large parameter $\beta>0$. 
In \eqref{eq:smooth_min}, the hyperparameter $c_Q>0$ dictates the trade-off between smoothness and accuracy of the approximation. With small enough $c_Q$, the smoothed threshold $\hat{s}_\text{th}^t$ recovers the original value $s_\text{th}^t$,
in the sense that we have the limit  $\lim_{c_Q \rightarrow 0}\hat{s}_{\text{th}}^t = s_\text{th}^t$.

Based on the differentiable approximation $\hat{s}^t_\text{th}$ introduced above, we can approximate the cardinality $|\Gamma(\mv x^t)|$ as the sum $\sum_{c=1}^C\mathbbm{1} (s_c(\mv x^{t}) \leq   \hat{s}_\text{th}^t  )$. Since the indicator function $\mathbbm{1}(\cdot)$ is also not differentiable, we replace the indicator function with the sigmoid function $\sigma(x)$ to obtain the following final differentiable approximation of the size of the set predictor
\begin{align}
   |\hat{\Gamma}(\mv x^{t})|=\sum_{c=1}^C\sigma\big( \hat{s}_\text{th}^t - s_c(\mv x^{t})\big).
	\label{asize}
\end{align}




\section{Experiments} \label{exp}
In this section, we provide experimental results to compare the performance of the adaptive point classifier DC-SNN \cite{li2023unleashing}, described in Sec.~\ref{sec:adaptive_point_classification},   and of the proposed set classifier SpikeCP. We also  provide insights into the trade-off between delay/energy and informativeness enabled by SpikeCP, as well as into the benefits of SpikeCP-based training. Finally, we offer a numerical comparison between the performance levels obtained by SpikeCP with Bonferroni and Simes corrections. All the experiments were run over a GPU server with single NVIDIA A100 card.

\subsection{Datasets}
We present experiments for the MNIST-DVS dataset \cite{serrano2015poker}, the DVS128 Gesture dataset \citep{amir2017low}, and the CIFAR-10 dataset.

The MNIST-DVS dataset contains labelled $26 \times 26$ spiking signals of duration $T=80$ samples. Each data point contains $26 \times 26 = 676$ spiking signals, which are recorded from a DVS camera that is shown moving handwritten digits from $``0"$ to $``9"$ on a screen. The data set contains $8,000$ training examples, as well as $2,000$ examples used for calibration and testing. The calibration data set $\mathcal{D}^\text{cal}$ is obtained by randomly sampling $|\mathcal{D}^\text{cal}|$ examples from the $2,000$ data points allocated for calibration and testing, with the rest used for testing (see, e.g., \cite{angelopoulos2022conformal}). We adopt a fully connected SNN with one hidden layer having $1,000$ neurons.

The DVS128 Gesture dataset collects videos from a DVS camera that is shown an actor performing one of $11$ different gestures under three different illumination conditions. We divide each time series into $T=80$ time intervals, integrating the discrete samples within each interval to obtain a continuous-valued time sample \citep{fang2021incorporating}. The dataset contains 1176 training data and 288 test data, from which $|\mathcal{D}^{\rm cal}|=50$ examples are randomly chosen to serve as calibration data. The SNN architecture is constructed using a convolutional layer, encompassing batch normalization and max-pooling layer,  as well as a fully-connected layer as described in \citep{fang2021incorporating}.

The CIFAR-10 dataset consists of 60,000 $32 \times 32$ color images that are divided into 10 classes, with 6000 images per class. There are 50,000 training images and 10,000 test images.  We use $|\mathcal{D}^{\rm cal}|=50$ calibration samples, which are obtained by randomly selecting 50 data points from the test set. We adopt a ResNet-18 architecture in which conventional neurons are replaced with SRM neurons \citep{fang2021incorporating}. Each example is repeatedly presented to the SNN for $T=80$ times.

\subsection{Setting}
All SNN models are trained via the surrogate gradient method as in \cite{9317803}. Except for the SpikeCP-based training, all the results reported in this section adopt a pre-trained SNN that is trained by assuming $\lambda=0$ and $\mathcal{T}_s=\{T\}$ as discussed in Sec \ref{spikecp+}. We average the performance  measures introduced in Sec.~\ref{subsec:measures} over $50$ different realizations of calibration and test data set. For SpikeCP, we assume the set of possible checkpoints as $\mathcal{T}_s = \{20, 40, 60, 80\}$, and use the global NC score (\ref{nc}) for SpikeCP, and we set the target set size  to $I_{\rm th}=3$, unless specified otherwise. For a fair comparison, we use the \emph{top-3 predictor} $\hat{\Gamma}(\mv x)$ for DC-SNN and SP-SNN after a final point prediction is made. The top-3 predictor $\hat{\Gamma}(\mv x)$ is constructed by including the top three predicted classes with the highest probabilities in $\mathbf{p}(\mv x^{T_s(\scalebox{0.8}{\mv x})})$ in \eqref{prob} (see, e.g., \cite{cresswell2024conformal}).

\begin{figure*}[htp]
	\centering
	\includegraphics[width=6.1in]{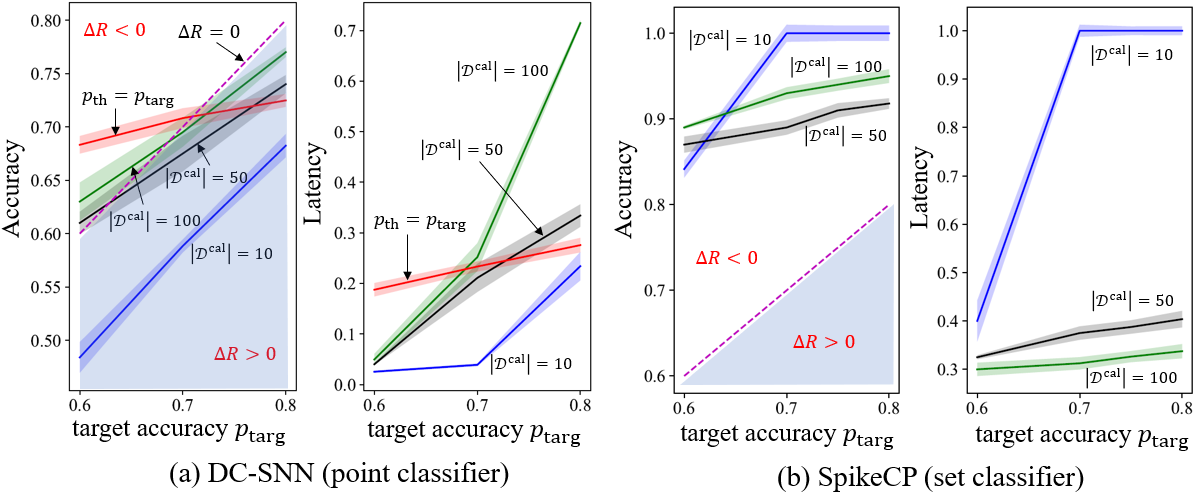}
    \caption{MNIST-DVS experiments: (a) Top-3 accuracy $\Pr(c \in \hat{\Gamma}(\mv x))$ and  normalized latency $\mathbb{E}[T_s(\mv x)]/T$ for the DC-SNN point classifier \citep{li2023unleashing}; (b) Accuracy $\Pr(c \in \Gamma(\mv x))$ and normalized  latency $\mathbb{E}[T_s(\mv x)]/T$ for the proposed SpikeCP set predictor given the target set size $I_{\rm th}=3$. The shaded error bars correspond to intervals covering $95\%$ of the realized values, obtained from $50$ different draws of calibration data.}
	\label{r1}
\end{figure*}

In this work, we implement the policy network of SP-SNN as a recurrent neural network (RNN) with one hidden layer having $500$ hidden neurons equipped with Tanh activation, followed by $T=80$ output neurons with a softmax activation function. The RNN takes the time series data $\mv x =\{\mv x_t\}_{t=1}^T$ as input, and outputs a probability vector $\mv{\pi}(\mv x|\mv \phi)$. The stopping time is chosen as $T_s(\mv x)=\arg\max_{t \in \{1,...,T\}}{\pi_t}(\mv x|\mv \phi)$ during the testing phase. The choice of a light-weight RNN architecture for policy network is dictated by the principle of ensuring that the size of the additional ANN is comparable to that of the SNN classifier \cite{li2023seenn}.

For SpikeCP-based training,  we assume data is split by considering the actual number of calibration data, i.e.,  $|\mathcal{D}^\text{tr,cal}|=\min\{|\mathcal{D}^\text{cal}|,|\mathcal{D}^\text{tr}|/2\}$, which also ensures a non-empty set $\mathcal{D}^\text{tr,te}$. The hyperparameters $c_Q$ and $\beta$ are set to $0.001$ and $1$, respectively. The weight factor $\lambda$ is set to 0.01, and the target accuracy level is set to $0.9$, i.e.,  $\alpha=0.1$.

\subsection{Performance Analysis of SpikeCP with a Pre-Trained SNN} \label{exp:spikecp}

We start by evaluating the performance with the same pre-trained SNN model for all schemes. Fig.~\ref{r1} reports accuracy -- $\Pr(c \in \hat{\Gamma}(\mv x))$ for DC-SNN and $\Pr(c \in \Gamma(\mv x))$ for SpikeCP -- and normalized latency $\mathbb{E}[T_s(\mv x)]/T$ as a function of the target accuracy $p_\mathrm{targ}$ for different sizes $|\mathcal{D}^\text{cal}|$ of the calibration data set on the MNIST-DVS dataset. The accuracy plots highlight the regime in which we have a  positive reliability gap $\Delta R$ in (\ref{reliap}) and (\ref{relia}), which corresponds to \emph{unreliable} decisions. 

For reference, in Fig.~\ref{r1}(a), we show the performance obtained by setting the threshold $p_\mathrm{th}$ in (\ref{stop}) to the accuracy target $p_\mathrm{targ}$. Following the results reported in Fig. \ref{fig:overall}(b), this approach yields unreliable decisions as soon as the target accuracy level is sufficiently large, here larger than $0.7$. By leveraging calibration data, DC-SNN can address this problem, suitably increasing the decision latency as $p_{\mathrm{targ}}$ increases. However, reliability -- i.e., a non-positive reliability gap -- is only approximately guaranteed when the number of calibration data points is sufficiently large, here $|\mathcal{D}^\text{cal}|=100$. 

In contrast, as shown in Fig.~\ref{r1}(b) and proved in Theorem 1, SpikeCP is always reliable, achieving a non-positive reliability gap irrespective of the number of calibration data points. With a fixed threshold $I_{\rm th}$, as in this example, increasing the size $|\mathcal{D}^\text{cal}|$ of the calibration data set has the effect of significantly reducing the average latency.

\begin{figure}
\centering
\includegraphics[width=2.8in]{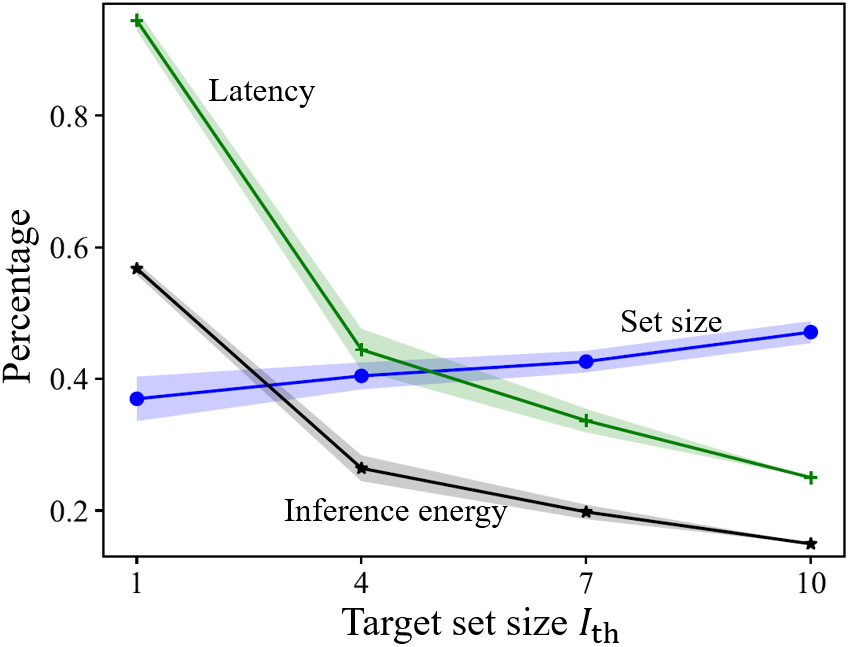}
    \caption{MNIST-DVS experiments: Normalized latency, inference energy, and set size (informativeness) as a function of target set size $I_{\rm th}$ for SpikeCP, assuming $p_{\rm targ}=0.9$ and $|\mathcal{D}^{\rm cal}|=200$ under the same conditions as Fig. 4. }
\label{r2}
\end{figure}

The trade-off supported by SpikeCP between latency and energy, on the one hand, and informativeness, i.e., set size, on the other hand,  is investigated in Fig.~\ref{r2} by varying the target set size $I_\text{th}$, with target accuracy level $p_\text{targ}=0.9$ and $|\mathcal{D}^\text{cal}|=200$ calibration examples on the MNIST-DVS dataset. Note that the reliability gap is always negative as in Fig.~\ref{r1}(b), and is hence omitted in the figure to avoid clutter. Increasing the target set size, $I_{\rm th}$, causes the final predicted set size, shown in the figure normalized by the number of classes $C=10$, to increase, yielding less informative decisions. On the flip side, sacrificing informativeness entails a lower (normalized) latency, as well as, correspondingly, a lower inference energy, with the latter shown in the figure as the average number of spikes per sample and per hidden neuron, $\mathbb{E}[S(\mv x)]/(1000T)$.

\begin{figure}[t!]
	\centering	\includegraphics[width=3.5in]{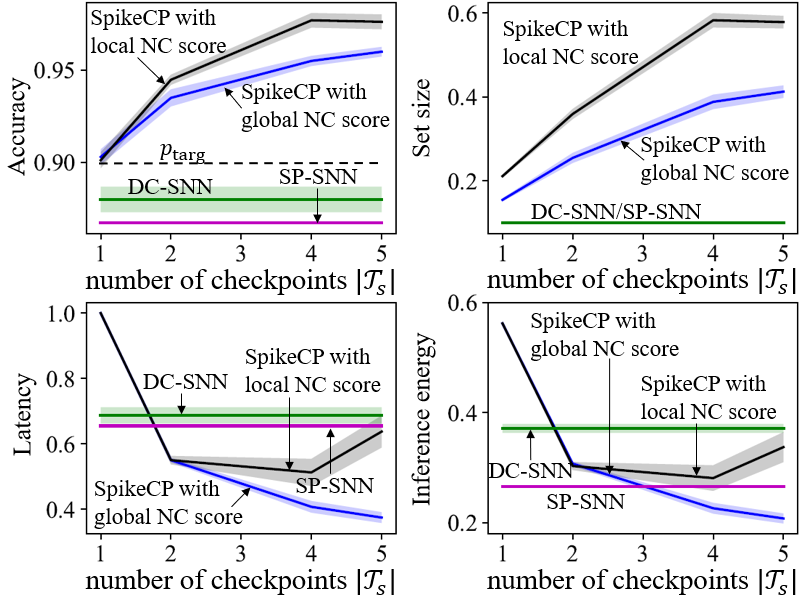}
    \caption{MNIST-DVS experiments: Accuracy, normalized latency, normalized set size (informativeness), and normalized inference energy as a function of number of checkpoints $|\mathcal{T}_s|$ for SpikeCP with local and global scores, as well as for DC-SNN and SP-SNN point classifiers, with  $p_{\rm targ}=0.9$, $|\mathcal{D}^{\rm cal}|=200$, and $I_{\rm th}=3$.  }
	\label{r3}
\end{figure}

In Fig.~\ref{r3}, we show the performance of SpikeCP when using either local NC scores (\ref{ncs}) or global NC scores (\ref{nc}) (see Sec.~\ref{spikecp}), as well as the performance of the DC-SNN and SP-SNN point predictors, as a function of the number of checkpoints $|\mathcal{T}_s|$, for $p_{\rm targ}=0.9$, $|\mathcal{D}^\text{cal}|=200$, and $I_\mathrm{th}=3$ on the MNIST-DVS dataset. The checkpoints are equally spaced among the $T$ time steps, and hence the checkpoint set is  $\mathcal{T}_s=\{T/|\mathcal{T}_s|, 2T/|\mathcal{T}_s|, ..., T\}$. The metrics displayed in the four panels are the accuracy  -- probability $\Pr(c \in \hat{\Gamma}(\mv x))$ for point predictors and probability $\Pr(c \in \Gamma(\mv x))$ for set predictors -- along with  normalized latency $\mathbb{E}[T_s(\mv x)]/T$ and normalized, per-neuron and per-time step,  inference energy  $\mathbb{E}[S(\mv x)]/(1000T)$. Note that the operation of SP-SNN and DC-SNN does not depend on the number of checkpoints, and hence the performance of these schemes is presented as a constant function.

By Theorem 1, SpikeCP always achieves negative reliability gap, while SP-SNN and DC-SNN fall short of the target reliability $p_{\text{targ}}$ in this example. Using global NC scores with SpikeCP yields better performance in terms of informativeness, i.e., set size, as well as latency and inference energy. The performance gap between the two choices of NC scores increases with the number of checkpoints, demonstrating that local NC scores are more sensitive to the Bonferroni correction applied by SpikeCP (see Sec. 4). This is due to the lower discriminative power of local confidence levels, which yield less informative NC scores (see, e.g., \cite{fisch2020efficient}). That said, moderate values of latency and inference energy can also be obtained with local NC scores, without requiring any coordination among the readout neurons. This can be considered to be one of the advantages of the calibration afforded by the use of SpikeCP.

With global NC scores, the number of checkpoints $|\mathcal{T}_s|$ is seen to control the trade-off between latency and informativeness for SpikeCP. In fact, a larger number of checkpoints improves the resolution of the stopping times, while at the same time yielding more conservative set-valued decision at each time step due to the mentioned Bonferroni correction. 

\begin{figure}[t!]
	\centering	\includegraphics[width=3.5in]{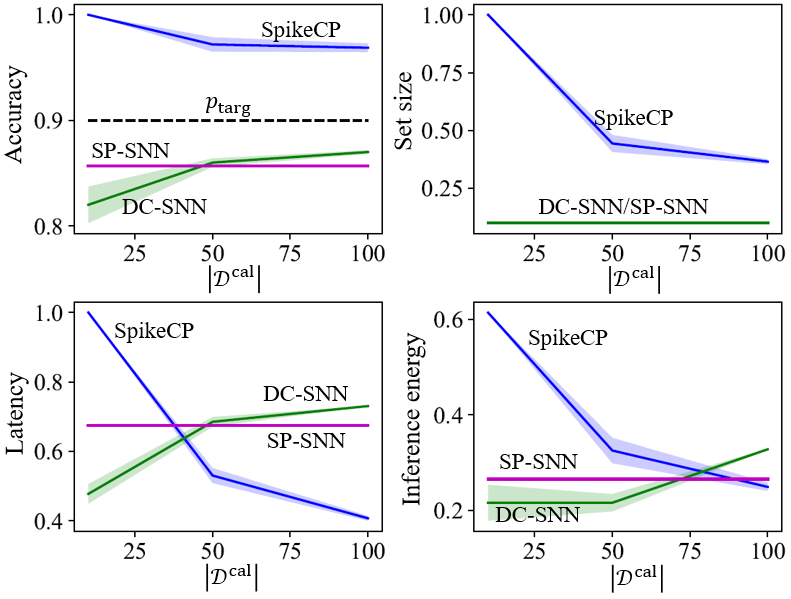}
    \caption{MNIST-DVS experiments: Accuracy, normalized latency, normalized set size (informativeness), and normalized inference energy as a function of number $|\mathcal{D}^\text{cal}|$ of calibration data points for SpikeCP with global NC scores, as well as for DC-SNN and SP-SNN point classifiers, with $p_{\rm targ}=0.9$, $|\mathcal{T}_s|=4$, and $I_{\rm th}=3$.  }
	\label{r4}
\end{figure} 

\begin{figure*}[htp]
	\centering
	\includegraphics[width=6.1in]{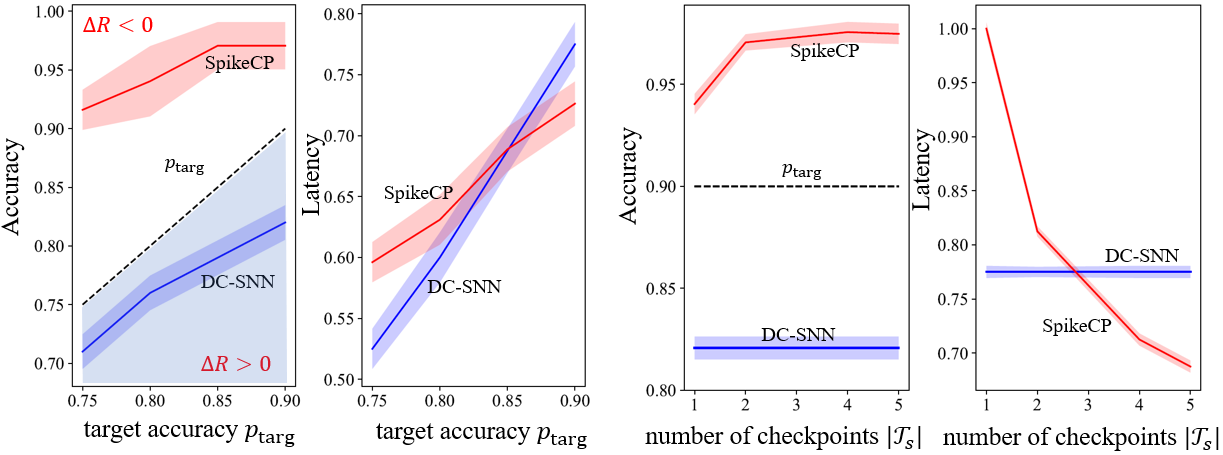}
    \caption{DVS128 Gesture experiments: Top-3 accuracy ($\Pr(c \in \hat{\Gamma}(\mv x))$ for DC-SNN and $\Pr(c \in \Gamma(\mv x))$ for SpikeCP), and normalized latency $\mathbb{E}[T_s(\mv x)]/T$ for the proposed SpikeCP set predictor and DC-SNN as a function of target accuracy $p_{\rm targ}$ and the number of checkpoints $|\mathcal{T}_s|$ with $|\mathcal{D}^{\rm cal}|=50$, $p_{\rm targ}=0.9$, $|\mathcal{T}_s|=4$ and $I_{\rm th}=3$.}
	\label{gesture}
\end{figure*} 

In Fig.~\ref{r4}, we show the performance of SpikeCP with global NC scores, DC-SNN, and SP-SNN as a function of the number, $|\mathcal{D}^{\rm cal}|$, of calibration data points, with $p_{\rm targ}=0.9$,  $|\mathcal{T}_s|=4$, and $I_\mathrm{th}=3$ on the MNIST-DVS dataset. The general conclusions around the comparisons among the different schemes are aligned with those presented above for Fig.~\ref{r3}. The figure also reveals that SP-SNN outperforms DC-SNN when the calibration data set is small, while DC-SNN is preferable in the presence of a sufficiently large data set. Finally, with a larger calibration data set, SpikeCP is able to increase the informativeness of the predicted set, while also decreasing latency and inference energy. 

Fig.~\ref{gesture} reports the accuracy and latency for DC-SNN and SpikeCP with different target accuracy $p_{\rm targ}$ and number of checkpoints $|\mathcal{T}_s|$ on the DVS128 Gesture dataset. In a manner consistent with the MNIST-DVS results, DC-SNN fails to meet the target accuracy, despite increasing the latency as $p_{\rm targ}$ increases. In contrast, SpikeCP is reliable, providing a negative reliability gap for all values $p_{\rm targ}$.  Furthermore, increasing the number of checkpoints, $|\mathcal{T}_s|$, the latency of SpikeCP decreases, since the SNN has more, earlier,  choices of times at which to stop inference. 

\begin{figure*}[t!]
	\centering
	\includegraphics[width=6.1in]{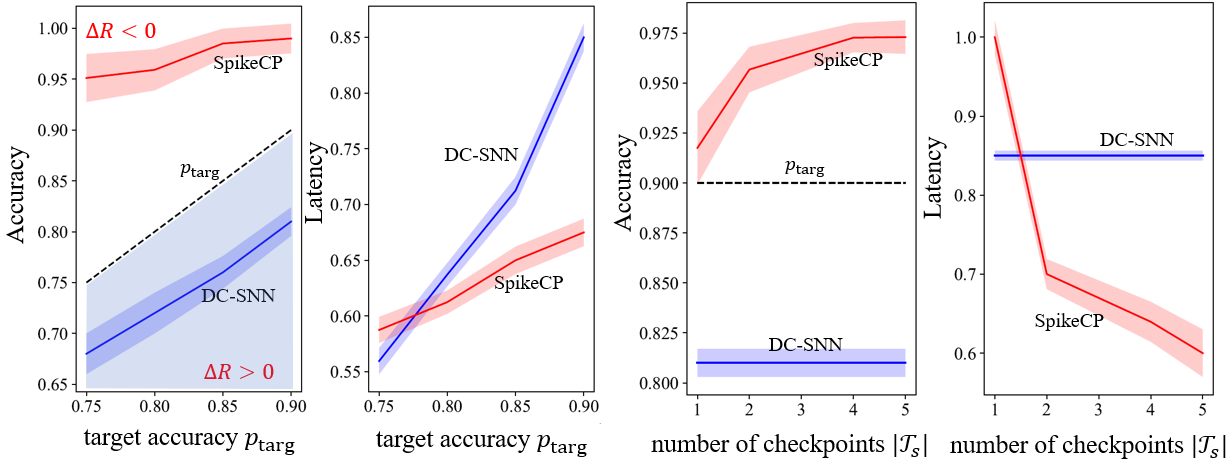}
    \caption{CIFAR-10 experiments: Top-3 accuracy ($\Pr(c \in \hat{\Gamma}(\mv x))$ for DC-SNN and $\Pr(c \in \Gamma(\mv x))$ for SpikeCP), and normalized latency $\mathbb{E}[T_s(\mv x)]/T$ for the proposed SpikeCP set predictor and DC-SNN as a function of target accuracy $p_{\rm targ}$ with $|\mathcal{D}^{\rm cal}|=50$, $p_{\rm targ}=0.9$, $|\mathcal{T}_s|=4$ and $I_{\rm th}=3$.} 
	\label{cifar10}
\end{figure*} 

In Fig.~\ref{cifar10}, we demonstrate the accuracy and normalized latency of SpikeCP and DC-SNN as a function of the target accuracy $p_{\rm targ}$ and of the number of checkpoints $|\mathcal{T}_s|$ on the CIFAR-10 dataset. The general conclusions reached from the analysis of these results are aligned with the insights obtained from the experiments reported on the MNIST-DVS dataset and DVS128 Gesture dataset. In particular, SpikeCP is seen to guarantee reliability regardless of the target accuracy and of the number of checkpoints, while DC-SNN cannot meet the target accuracy. Furthermore,  the inference latency decreases with a larger  number of checkpoints due to the larger granularity of the stopping times allowed for SpikeCP and due to the larger Bonferroni correction imposed for each checkpoint time.

\subsection{Comparing Bonferroni and Simes Corrections} \label{exp:bs}

\begin{figure}[htp]
	\centering	\includegraphics[width=3.5in]{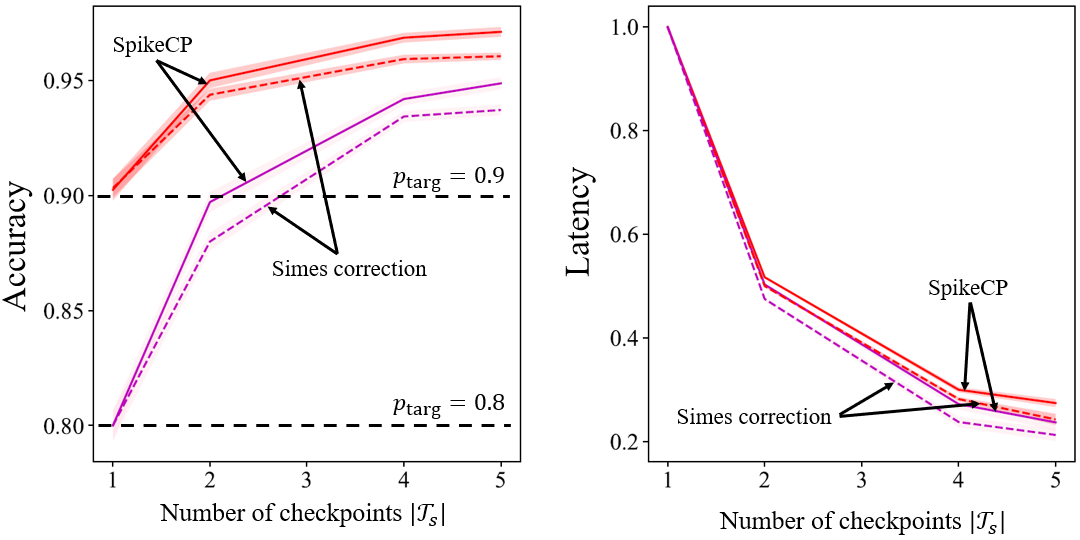}
    \caption{Accuracy and normalized latency as a function of number of checkpoints $|\mathcal{T}_s|$ for SpikeCP, which uses the Bonferroni correction, as well as for a variant that applies Simes correction (see Sec. \ref{sec:simes}), with $p_{\rm targ}=0.8$ and $p_{\rm targ}=0.9$ on the MNIST-DVS dataset. }
	\label{r7}
\end{figure} 

In Fig.~\ref{r7}, we study the performance of SpikeCP, which uses  Bonferroni correction  (see Sec. \ref{eth}), with a heuristic variant of SpikeCP that uses Simes correction (see Sec. \ref{sec:simes}) with $p_{\rm targ}=0.8$ and $p_{\rm targ}=0.9$. Fig. \ref{r7} plots accuracy and normalized latency as a function of the number of checkpoints, on the MNIST-DVS dataset. As discussed in Sec. \ref{sec:simes}, the Bonferroni correction applied by SpikeCP becomes increasingly strict as the number of checkpoints increases. Accordingly, alternative correction factors, such as Simes, may become advantageous in the regime of large number of checkpoints. Confirming this argument, the figures show that indeed Simes correction can yield some advantage in terms of latency, while still satisfying, despite its lack of theoretical guarantees, the reliability requirement \eqref{reliap}. 

\subsection{Performance Analysis of SpikeCP-based Training}
We finally turn to analyzing the potential benefits of SikeCP-based training, as introduced in Sec.~\ref{spikecp+}. Accordingly, the SNN classifier is trained by minimizing the objective in \eqref{oloss}, with hyperparameter $\lambda$ dictating the relative weight given to the prediction set efficiency over the conventional cross-entropy performance metric. With $\lambda=0$, we recover the same SNN model assumed throughout the rest of the section, while larger values of $\lambda>0$ ensure that the training model is increasingly tailored to the use of SpikeCP during inference by targeting the predictive set inefficiency. 

\begin{figure}[t!]
\centering
\includegraphics[width=3.1in]{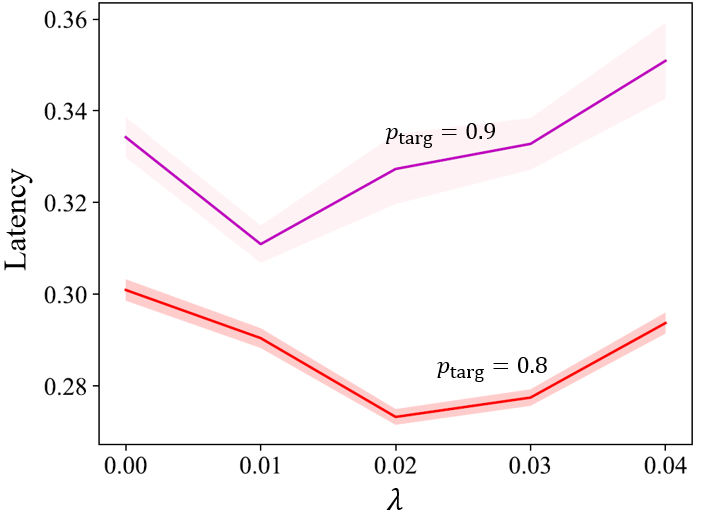}
    \caption{Normalized latency as a function of the weight factor $\lambda$ in the training objective \eqref{oloss} for training-based SpikeCP under target accuracy $p_{\rm targ}=0.8$ and $p_{\rm targ}=0.9$, assuming $|\mathcal{D}^{\rm cal}|=200$ calibration data points with the same other conditions as in Fig. 4, on the MNIST-DVS dataset. }
\label{r5}
\end{figure}

In order to elaborate on the choice of hyperparameter $\lambda$, in Fig.~\ref{r5} we plot the normalized latency of SpikeCP as a function of $\lambda$ on the MNIST-DVS dataset. For both target accuracy values $p_{\rm targ}=0.8$ and $p_{\rm targ}=0.9$, it is observed that there is an optimal value of $\lambda$ that balances the inefficiency and accuracy (cross-entropy) criteria. Increasing $\lambda$  is initially beneficial, yielding smaller predictive sets and hence smaller latencies. However, larger values of $\lambda$ eventually downweigh excessively the accuracy criterion, producing worse performance. Furthermore, the optimal value of $\lambda$ is seen to be decreasing with growing target reliability levels $p_{\rm targ}$, which call for more emphasis on the cross-entropy criterion.
 
\begin{figure}
\centering
\includegraphics[width=3.5in]{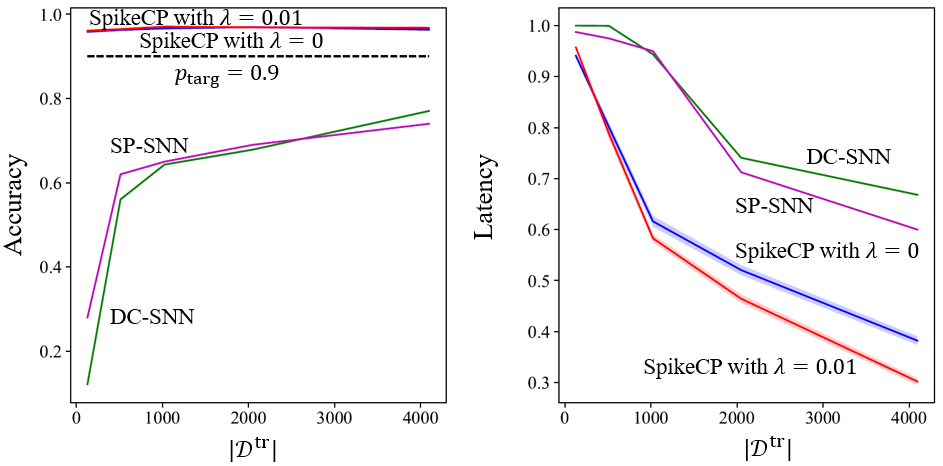}
    \caption{Accuracy and normalized latency as a function of the number of training data $|\mathcal{D}^{\rm tr}|$, assuming $p_{\rm targ}=0.9$, and $|\mathcal{D}^{\rm cal}|=100$ under the same conditions as Fig. 4, on the MNIST-DVS dataset. }
\label{r6}
\end{figure}

We now turn to comparing the performance of SpikeCP-based training with conventional SpikeCP (with $\lambda=0$), DC-SNN and SP-SNN. Specifically, Fig.~\ref{r6} plots accuracy and normalized latency as a function of the number of training data points $|\mathcal{D}^{\rm tr}|$, on the MNIST-DVS dataset. The point classifiers DC-SNN and SP-SNN exhibit an increasing accuracy level as the  training data set size increases, while still failing to meet the reliability target $p_{\textrm{targ}}=0.9$. In contrast, SpikeCP schemes meet the reliability requirement for any number of training data points. More training data translate into a lower latency, with SpikeCP-based training, here run with $\lambda=0.01$, proving an increasingly sizeable latency reduction.

\section{Conclusions} \label{conclusion}
In this work, we have introduced SpikeCP, a delay-adaptive SNN set predictor with provable reliability guarantees. SpikeCP wraps around any pre-trained SNN classifier, producing a set classifier with a tunable trade-off between informativeness of the decision -- i.e., size of the predicted set -- and latency, or inference energy as measured by the number of spikes. Unlike prior art, the reliability guarantees of SpikeCP hold irrespective of the quality of the pre-trained SNN and of the number of calibration points, with minimal added complexity. SpikeCP was also integrated with a CP-aware training strategy that complements the conventional cross-entropy criterion with a regularizer accounting for the informativeness of the predicted set. 

Among the possible extensions of this study, one could consider regression problems. In this case, the spike count $r(\mv x^t)$ produced by an output neuron could be converted into a scalar decision $\hat{c}(\mv x^t)$, e.g., via rate or temporal decoding \cite{jang2019introduction}. SpikeCP can be extended to address such a situation by adopting an NC score given by a regression loss like $s_c(\mv x^t)=|c-\hat{c}(\mv x^t)|$, where $c$ is the real-valued scalar being predicted; as well as by modifying the stopping criterion \eqref{eq:thresholdset} so that the set size $|\Gamma(\mv x^t)|$ represents the size of the predicted interval. 

For another future work, we highlight extensions of SpikeCP that take into account time decoding or Bayesian learning \cite{jang2021bisnn} in order to further reduce the number of spikes and enhance the reliability of confidence estimates.

\section*{Appendix: CP and Hypothesis Testing} \label{app:CP_SHT}

As detailed in Sec. \ref{spikecp}, SpikeCP relies on the use of the Bonferroni, or Simes, corrections, which are tools introduced in the literature on hypothesis testing \cite{shaffer1995multiple}.  In this appendix, we elaborate on the connection between CP and multiple-hypothesis testing. 


Conventional CP effectively applies a binary hypothesis test for each possible label $c$, testing the null hypothesis that the label $c$ is the correct one. With the notation of this paper, for any fixed time $t$, CP considers the null hypothesis
\begin{align*}
    \mathcal{H}_t(\boldsymbol{x}^t,c): \text{$(\boldsymbol{x}^t, c)$ and the calibration data $\mathcal{D}^{t, \rm cal}$} ~\text{are i.i.d.},
\end{align*}
where we have defined $\mathcal{D}^{t, \rm cal}=\{(\mv x^t[i], c[i])\}_{i=1}^{|\mathcal{D}^{\rm cal}|}$. In fact, if this hypothesis holds  true, label $c$ is the ground-truth label for input $\boldsymbol{x}^t$. 

Suppose that we have a valid $p$-variable $ p_t(\boldsymbol{x}^t,c)$ for this hypothesis, i.e., a random variable -- which may be also a function of the calibration data -- that satisfies the inequality $\Pr(p_t(\boldsymbol{x}^t,c) \leq  \alpha'|\mathcal{H}_t(\boldsymbol{x}^t,c))\leq \alpha'$ for all $\alpha' \in [0,1]$, where the probability is conditioned over the hypothesis being correct. Then, constructing the predictive set as $\Gamma(\boldsymbol{x}^t)=\{c\in\mathcal{C}: p_t(\boldsymbol{x}^t,c) > \alpha'\}$ would guarantee the reliability condition $\Pr\big(c\in \Gamma(\mv x^{t} )\big)\geq 1-\alpha'$ for the given fixed time $t$.

The key underlying technical result in the theory of CP is that the variable \begin{align}
    \label{eq:conformal_p_value}
    p_t(\mv x^t,c) = \frac{ 1+\sum_{i=1}^{|\mathcal{D}^\text{cal}|}\mathbbm{1}(s_{c}(\mv x^t) \leq s_{c[i]}(\mv x^t[i]) )}{|\mathcal{D}^\text{cal}|+1},
\end{align}
is a valid $p$-variable for time $t$, where $s_{c}(\mv x^t)$ is an NC score. 
The predictive set constructed by $p$-value $\Gamma(\boldsymbol{x}^t)=\{c\in\mathcal{C}: p_t(\boldsymbol{x}^t,c) > \alpha\}$ is equivalent to the expression of \eqref{cpsett} since \emph{excluding} the $\alpha$-fraction ($p_t(\mv x^t,c)>\alpha$) is equivalent to \emph{including} the $(1-\alpha)$-fraction $(s_c(\mv x^t) \leq s_\text{th}^t)$. 

In SpikeCP, the time $t=T_s(\mv x)$ at which a decision is made depends on the input $\mv x$, and hence the reliability guarantees described above do not apply directly. What is needed, instead, are \emph{corrected} $p$-variables $\tilde{p}_t(\mv x^t,c)$ satisfying the property $\Pr\big(\tilde{p}_t(\boldsymbol{x}^t,c) > \alpha' \textrm{ for all $t\in\mathcal{T}_s$}|\mathcal{H}(\boldsymbol{x},c)\big)\geq 1-\alpha'$ for all $\alpha' \in [0,1]$, where, under the composite null hypothesis $\mathcal{H}(\boldsymbol{x},c)$, the pair $(\boldsymbol{x}, c)$ and the calibration data $\mathcal{D}^{\rm cal}$ are i.i.d. Note that the hypothesis $\mathcal{H}(\boldsymbol{x},c)$ implies all hypotheses $\mathcal{H}_t(\boldsymbol{x}^t,c)$ for $t\in\mathcal{T}_s$.

To find such corrected $p$-variables, it is sufficient to identify a valid $p$-variable $p(\mv x,c)$ for the composite hypothesis $\mathcal{H}(\boldsymbol{x},c)$ such that, with probability 1, we have the inequalities $\tilde{p}_t(\mv x^t,c) \geq p(\mv x,c)$ for suitable functions $\tilde{p}_t(\mv x^t,c)$ of the original $p$-values ${p}_t(\mv x^t,c)$. Bonferroni's method provides one such $p$-variable, namely $p^{\text{B}}(\mv x, c) = \min_{t \in \mathcal{T}_s}\{ |\mathcal{T}_s| p_t(\mv x^t, c) \}$ with corrected $p$-variables $\tilde{p}_t(\mv x^t,c) = |\mathcal{T}_s|p_t(\mv x^t,c)$ \cite[Appendix 2]{hochberg1987multiple}. It can be checked that this selection yields the SpikeCP procedure in Algorithm 1.

Alternatively, the composite $p$-value produced by Simes correction is $p^\text{S}(\mv x,c) = \min_{t \in \mathcal{T}_s}\{  {|\mathcal{T}_s|} p_t(\mv x^t,c)/{r(t)}  \}$, where $r(t)$ is the ranking of $p_t(\mv x^t, c)$ among $\{p_t(\mv x^t, c)\}_{t \in \mathcal{T}_s}$.  This yields corrected $p$-variables $\tilde{p}_t(\mv x^t,c) = |\mathcal{T}_s|p_t(\mv x^t,c)/{r(t)}$. Reference \cite{rodland2006simes} proved that this approach provides a valid $p$-value as long as the joint distribution over the $|\mathcal{T}_s|$ $p$-values ${p}_t(\mv x^t,c)$ have the \emph{multivariate totally positive of order $2$} $(\text{MTP}_2)$ property as defined in \cite{rodland2006simes}. Together with the assumption of increasing $p$-value, Simes corrected $p$-values yield the heuristic SpikeCP variant discussed in Sec. \ref{sec:simes}.

\small{
\bibliographystyle{IEEEtran}
\bibliography{references}}

\begin{thebibliography}{10}
\providecommand{\url}[1]{#1}
\csname url@samestyle\endcsname
\providecommand{\newblock}{\relax}
\providecommand{\bibinfo}[2]{#2}
\providecommand{\BIBentrySTDinterwordspacing}{\spaceskip=0pt\relax}
\providecommand{\BIBentryALTinterwordstretchfactor}{4}
\providecommand{\BIBentryALTinterwordspacing}{\spaceskip=\fontdimen2\font plus
\BIBentryALTinterwordstretchfactor\fontdimen3\font minus
  \fontdimen4\font\relax}
\providecommand{\BIBforeignlanguage}[2]{{%
\expandafter\ifx\csname l@#1\endcsname\relax
\typeout{** WARNING: IEEEtran.bst: No hyphenation pattern has been}%
\typeout{** loaded for the language `#1'. Using the pattern for}%
\typeout{** the default language instead.}%
\else
\language=\csname l@#1\endcsname
\fi
#2}}
\providecommand{\BIBdecl}{\relax}
\BIBdecl

\bibitem{davies2018loihi}
M.~Davies \emph{et~al.}, ``Loihi: A neuromorphic manycore processor with
  on-chip learning,'' \emph{IEEE Micro}, vol.~38, no.~1, pp. 82--99, 2018.

\bibitem{li2023unleashing}
C.~Li, E.~G. Jones, and S.~Furber, ``Unleashing the potential of spiking neural
  networks with dynamic confidence,'' in \emph{Proc. IEEE/CVF International
  Conference on Computer Vision}, 2023, pp. 13\,350--13\,360.

\bibitem{li2023seenn}
Y.~Li, T.~Geller, Y.~Kim, and P.~Panda, ``{SEENN}: Towards temporal spiking
  early-exit neural networks,'' \emph{arXiv preprint arXiv:2304.01230}, 2023.

\bibitem{li2021free}
Y.~Li, S.~Deng, X.~Dong, R.~Gong, and S.~Gu, ``A free lunch from {ANN}: Towards
  efficient, accurate spiking neural networks calibration,'' in \emph{ICML},
  pp. 6316--6325, 2021.

\bibitem{serrano2015poker}
T.~Serrano-Gotarredona and B.~Linares-Barranco, ``Poker-{DVS} and {MNIST-DVS}.
  their history, how they were made, and other details,'' \emph{Frontiers in
  Neuroscience}, vol.~9, p. 481, 2015.

\bibitem{vovk2022algorithmic}
V.~Vovk, A.~Gammerman, and G.~Shafer, \emph{Algorithmic {L}earning in a
  {R}andom {W}orld}.\hskip 1em plus 0.5em minus 0.4em\relax Springer Nature,
  2022.

\bibitem{angelopoulos2107gentle}
A.~N. Angelopoulos and S.~Bates, ``A gentle introduction to conformal
  prediction and distribution-free uncertainty quantification. arxiv 2021,''
  \emph{arXiv preprint arXiv:2107.07511}, 2021.

\bibitem{fisch2020efficient}
A.~Fisch, T.~Schuster, T.~Jaakkola, and R.~Barzilay, ``Efficient conformal
  prediction via cascaded inference with expanded admission,'' \emph{arXiv
  preprint arXiv:2007.03114}, 2020.

\bibitem{rodland2006simes}
E.~A. R{\o}dland, ``Simes' procedure is ‘valid on average’,''
  \emph{Biometrika}, vol.~93, no.~3, pp. 742--746, 2006.

\bibitem{wade2010swat}
J.~J. Wade, L.~J. McDaid, J.~A. Santos, and H.~M. Sayers, ``{SWAT}: A spiking
  neural network training algorithm for classification problems,'' \emph{IEEE
  Transactions on Neural Networks}, vol.~21, no.~11, pp. 1817--1830, 2010.

\bibitem{caporale2008spike}
N.~Caporale and Y.~Dan, ``Spike timing--dependent plasticity: a {Hebbian}
  learning rule,'' \emph{Annu. Rev. Neurosci.}, vol.~31, pp. 25--46, 2008.

\bibitem{neftci2019surrogate}
E.~O. Neftci, H.~Mostafa, and F.~Zenke, ``Surrogate gradient learning in
  spiking neural networks: Bringing the power of gradient-based optimization to
  spiking neural networks,'' \emph{IEEE Signal Processing Magazine}, vol.~36,
  no.~6, pp. 51--63, 2019.

\bibitem{jang2019introduction}
H.~Jang, O.~Simeone, B.~Gardner, and A.~Gruning, ``An introduction to
  probabilistic spiking neural networks: Probabilistic models, learning rules,
  and applications,'' \emph{IEEE Signal Processing Magazine}, vol.~36, no.~6,
  pp. 64--77, 2019.

\bibitem{skatchkovsky2022bayesian}
N.~Skatchkovsky, H.~Jang, and O.~Simeone, ``Bayesian continual learning via
  spiking neural networks,'' \emph{arXiv preprint arXiv:2208.13723}, 2022.

\bibitem{guo2017calibration}
C.~Guo \emph{et~al.}, ``On calibration of modern neural networks,'' in
  \emph{ICML}, pp. 1321--1330, 2017.

\bibitem{quach2023conformal}
V.~Quach, A.~Fisch, T.~Schuster, A.~Yala, J.~H. Sohn, T.~S. Jaakkola, and
  R.~Barzilay, ``Conformal language modeling,'' \emph{arXiv preprint
  arXiv:2306.10193}, 2023.

\bibitem{rosenfeld2019learning}
B.~Rosenfeld, O.~Simeone, and B.~Rajendran, ``Learning first-to-spike policies
  for neuromorphic control using policy gradients,'' in \emph{Proc. IEEE
  SPAWC}, pp. 1--5, 2019.

\bibitem{panda2016conditional}
P.~Panda, A.~Sengupta, and K.~Roy, ``Conditional deep learning for
  energy-efficient and enhanced pattern recognition,'' in \emph{Proc. IEEE
  DATE}, pp. 475--480, 2016.

\bibitem{teerapittayanon2016branchynet}
S.~Teerapittayanon, B.~McDanel, and H.-T. Kung, ``Branchynet: Fast inference
  via early exiting from deep neural networks,'' in \emph{Proc. IEEE ICPR}, pp.
  2464--2469, 2016.

\bibitem{laskaridis2020spinn}
S.~Laskaridis, S.~I. Venieris, M.~Almeida, I.~Leontiadis, and N.~D. Lane,
  ``{SPINN}: synergistic progressive inference of neural networks over device
  and cloud,'' in \emph{Proc. MobiCom}, pp. 1--15, 2020.

\bibitem{pacheco2023impact}
R.~G. Pacheco, R.~S. Couto, and O.~Simeone, ``On the impact of deep neural
  network calibration on adaptive edge offloading for image classification,''
  \emph{Journal of Network and Computer Applications}, p. 103679, 2023.

\bibitem{weiss2010structured}
D.~Weiss and B.~Taskar, ``Structured prediction cascades,'' in \emph{Proc.
  AISTATS}, pp. 916--923, 2010.

\bibitem{cohen2022calibrating}
K.~M. Cohen, S.~Park, O.~Simeone, and S.~Shamai, ``Calibrating {AI} models for
  wireless communications via conformal prediction,'' \emph{arXiv preprint
  arXiv:2212.07775}, 2022.

\bibitem{lin2022conformal}
Z.~Lin, S.~Trivedi, and J.~Sun, ``Conformal prediction with temporal quantile
  adjustments,'' \emph{arXiv preprint arXiv:2205.09940}, 2022.

\bibitem{angelopoulos2022conformal}
A.~N. Angelopoulos \emph{et~al.}, ``Conformal risk control,'' \emph{arXiv
  preprint arXiv:2208.02814}, 2022.

\bibitem{kumar2023conformal}
B.~Kumar \emph{et~al.}, ``Conformal prediction with large language models for
  multi-choice question answering,'' \emph{arXiv preprint arXiv:2305.18404},
  2023.

\bibitem{stutz2021learning}
D.~Stutz \emph{et~al.}, ``Learning optimal conformal classifiers,'' \emph{arXiv
  preprint arXiv:2110.09192}, 2021.

\bibitem{einbinder2022training}
B.-S. Einbinder, Y.~Romano, M.~Sesia, and Y.~Zhou, ``Training uncertainty-aware
  classifiers with conformalized deep learning,'' \emph{arXiv preprint
  arXiv:2205.05878}, 2022.

\bibitem{park2022few}
S.~Park, K.~M. Cohen, and O.~Simeone, ``Few-shot calibration of set predictors
  via meta-learned cross-validation-based conformal prediction,'' \emph{IEEE
  Transactions on Pattern Analysis and Machine Intelligence}, vol.~46, no.~1,
  pp. 280--291, 2024.

\bibitem{fisch2021few}
A.~Fisch, T.~Schuster, T.~Jaakkola, and R.~Barzilay, ``Few-shot conformal
  prediction with auxiliary tasks,'' in \emph{ICML}, pp. 3329--3339, 2021.

\bibitem{vovk2022admissible}
V.~Vovk, B.~Wang, and R.~Wang, ``Admissible ways of merging p-values under
  arbitrary dependence,'' \emph{The Annals of Statistics}, vol.~50, no.~1, pp.
  351--375, 2022.

\bibitem{vovk2021values}
V.~Vovk and R.~Wang, ``E-values: Calibration, combination and applications,''
  \emph{The Annals of Statistics}, vol.~49, no.~3, pp. 1736--1754, 2021.

\bibitem{9997098}
J.~Chen, N.~Skatchkovsky, and O.~Simeone, ``Neuromorphic integrated sensing and
  communications,'' \emph{IEEE Wireless Communications Letters}, vol.~12,
  no.~3, pp. 476--480, 2023.

\bibitem{skatchkovsky2021spiking}
N.~Skatchkovsky, H.~Jang, and O.~Simeone, ``Spiking neural networks—{P}art
  {II}: Detecting spatio-temporal patterns,'' \emph{IEEE Communications
  Letters}, vol.~25, no.~6, pp. 1741--1745, 2021.

\bibitem{eshraghian2021training}
J.~K. Eshraghian \emph{et~al.}, ``Training spiking neural networks using
  lessons from deep learning,'' \emph{arXiv preprint arXiv:2109.12894}, 2021.

\bibitem{tao2023}
T.~Sun, B.~Yin, and S.~Bohte, ``Efficient uncertainty estimation in spiking
  neural networks via {MC}-dropout,'' \emph{arXiv preprint arXiv:2304.10191},
  2023.

\bibitem{9317803}
N.~Skatchkovsky, H.~Jang, and O.~Simeone, ``Spiking neural networks—{P}art
  {III}: Neuromorphic communications,'' \emph{IEEE Communications Letters},
  vol.~25, no.~6, pp. 1746--1750, 2021.

\bibitem{10016643}
J.~Chen, N.~Skatchkovsky, and O.~Simeone, ``Neuromorphic wireless cognition:
  Event-driven semantic communications for remote inference,'' \emph{IEEE
  Transactions on Cognitive Communications and Networking}, vol.~9, no.~2, pp.
  252--265, 2023.

\bibitem{cp2023}
\BIBentryALTinterwordspacing
R.~Tibshirani, ``Conformal prediction: Advanced topics in statistical learning
  (lecture note),'' 2023. [Online]. Available:
  \url{https://www.stat.berkeley.edu/~ryantibs/statlearn-s23/lectures/conformal.pdf}
\BIBentrySTDinterwordspacing

\bibitem{tibshirani2019conformal}
R.~J. Tibshirani, R.~Foygel~Barber, E.~Candes, and A.~Ramdas, ``Conformal
  prediction under covariate shift,'' \emph{Advances in neural information
  processing systems}, vol.~32, 2019.

\bibitem{kuchibhotla2020exchangeability}
A.~K. Kuchibhotla, ``Exchangeability, conformal prediction, and rank tests,''
  \emph{arXiv preprint arXiv:2005.06095}, 2020.

\bibitem{lei2018distribution}
J.~Lei \emph{et~al.}, ``Distribution-free predictive inference for
  regression,'' \emph{Journal of the American Statistical Association}, vol.
  113, no. 523, pp. 1094--1111, 2018.

\bibitem{amir2017low}
A.~Amir \emph{et~al.}, ``A low power, fully event-based gesture recognition
  system,'' in \emph{Proceedings of the IEEE conference on computer vision and
  pattern recognition}, 2017, pp. 7243--7252.

\bibitem{fang2021incorporating}
W.~Fang, Z.~Yu, Y.~Chen, T.~Masquelier, T.~Huang, and Y.~Tian, ``Incorporating
  learnable membrane time constant to enhance learning of spiking neural
  networks,'' in \emph{Proceedings of the IEEE/CVF international conference on
  computer vision}, 2021, pp. 2661--2671.

\bibitem{cresswell2024conformal}
J.~C. Cresswell, Y.~Sui, B.~Kumar, and N.~Vouitsis, ``Conformal prediction sets
  improve human decision making,'' 2024.

\bibitem{jang2021bisnn}
H.~Jang, N.~Skatchkovsky, and O.~Simeone, ``{BiSNN}: training spiking neural
  networks with binary weights via bayesian learning,'' in \emph{IEEE DSLW},
  pp. 1--6, 2021.

\bibitem{shaffer1995multiple}
J.~P. Shaffer, ``Multiple hypothesis testing,'' \emph{Annual review of
  psychology}, vol.~46, no.~1, pp. 561--584, 1995.

\bibitem{hochberg1987multiple}
Y.~Hochberg and A.~C. Tamhane, \emph{Multiple comparison procedures}.\hskip 1em
  plus 0.5em minus 0.4em\relax John Wiley \& Sons, Inc., 1987.

\end{thebibliography}

\end{document}